\documentclass{article}

\usepackage[preprint]{neurips_2025}

\usepackage[utf8]{inputenc} 
\usepackage[T1]{fontenc}    
\usepackage{hyperref}       
\usepackage{url}            
\usepackage{booktabs}       
\usepackage{amsfonts}       
\usepackage{nicefrac}       
\usepackage{microtype}      
\usepackage{xcolor}         

\title{Learning with Spike Synchrony in Spiking Neural Networks}

\author{%
  Yuchen Tian\textsuperscript{1}\thanks{Corresponding author: ytia0587@uni.sydney.edu.au},
  Assel Kembay\textsuperscript{2},
  Samuel Tensingh\textsuperscript{1},\\
  \textbf{Nhan Duy Truong\textsuperscript{1}}, 
  \textbf{Jason K.~Eshraghian\textsuperscript{2}},
  \textbf{Omid Kavehei\textsuperscript{1}} \\
  \\
  \textsuperscript{1}School of Biomedical Engineering, The University of Sydney, Sydney, NSW, Australia \\
  \textsuperscript{2}Department of Electrical and Computer Engineering, University of California, Santa Cruz, CA, USA 
}

\begin{document}

\maketitle


\begin{abstract}
Spiking neural networks (SNNs) promise energy-efficient computation by mimicking biological neural dynamics, yet existing plasticity rules focus on isolated spike pairs and fail to leverage the synchronous activity patterns that drive learning in biological systems. We introduce spike-synchrony-dependent plasticity (SSDP), a training approach that adjusts synaptic weights based on the degree of synchronous neural firing rather than spike timing order. Our method operates as a local, post-optimization mechanism that applies updates to sparse parameter subsets, maintaining computational efficiency with linear scaling. SSDP serves as a lightweight event-structure regularizer, biasing the network toward biologically plausible spatio-temporal synchrony while preserving standard convergence behavior. SSDP seamlessly integrates with standard backpropagation while preserving the forward computation graph. We validate our approach across single-layer SNNs and spiking Transformers on datasets from static images to high-temporal-resolution tasks, demonstrating improved convergence stability and enhanced robustness to spike-time jitter and event noise. 
These findings provide new insights into how biological neural networks might leverage synchronous activity for efficient information processing and suggest that synchrony-dependent plasticity represents a key computational principle underlying neural learning.
\end{abstract}


\section{Introduction}

The brain processes information through sparse, event-driven spikes, which allows it to perform robust, energy-efficient calculations in changing environments \cite{stuijt2021mubrain,malyshev2013energy,lewicki2002efficient,olshausen1996emergence}. Inspired by this biological principle, spiking neural networks (SNNs) \cite{maass1997networks,zenke2021remarkable,tavanaei2019deep,sharifshazileh2021electronic} imitate the spike-based communication used by real neurons. Unlike traditional artificial neural networks (ANNs), which rely on continuous-valued signals, SNNs use precise binary spikes that occur at specific times, forming a fundamentally different way of computing \cite{gutig2006tempotron}. This brain-inspired architecture supports sparse coding and asynchronous updates, making SNNs naturally energy-efficient and well-suited for neuromorphic hardware implementation \cite{indiveri2015memory}. However, training SNNs remains challenging because spikes are discontinuous and non-differentiable, often resulting in lower performance compared to standard deep learning models \cite{neftci2019surrogate, eshraghian2023training, pfeiffer2018deep}.

To address the limitations of gradient-based methods in SNNs, researchers have turned to biologically inspired synaptic plasticity rules. Among these, Hebbian learning first introduced the fundamental idea that correlated neural activity strengthens connections \cite{hebblearning}. Despite its biological relevance, classical Hebbian plasticity overlooks the precise temporal structure of spikes and lacks sensitivity to causality in spike timing. Spike-timing-dependent plasticity (STDP) was first experimentally observed by Bi and Poo \cite{bi1998synaptic}, who reported that the relative timing between pre- and post-synaptic spikes modulates synaptic strength. Although STDP is usually described using smooth exponential curves, the original experimental data were quite noisy, meaning these curves represent approximate trends rather than exact biological rules. Nevertheless, STDP has become widely adopted as a temporal learning mechanism in SNNs due to its biological inspiration and simplicity. It has been applied to various vision and classification tasks \cite{liu2021sstdp, goupy2024paired, goupy2024neuronal}. 

Reward-modulated STDP (R-STDP) \cite{quintana2022bio} incorporates global reward signals (analogous to dopamine) to gate synaptic updates, achieving biological plausibility but limited by coarse and delayed reinforcement signals. Supervised STDP approaches, such as SSTDP \cite{liu2021sstdp} and S2-STDP \cite{goupy2024neuronal}, rely on precise, externally provided spike-time errors, enabling fine-grained synaptic adjustments at the expense of biological realism due to their dependence on supervised signals. More recently, S-TLLR \cite{apolinario2023s} introduced a generalized three-factor eligibility trace that captures both causal and non-causal spike-time interactions, achieving good scalability and lower computational demands. However, STDP-style rules face two main limitations. Firstly, it generally focuses only on individual synapses and pairs of spikes, ignoring more complex collective spiking patterns. Second, its sensitivity to exact spike timing makes it vulnerable to noise and instability in high-frequency or dynamic environments \cite{popovych2013self}. The data by which STDP is derived is considerably noisier than the exponential curve fits. 

From an engineering standpoint, order-based, time-dependent rules also impose nontrivial time and memory complexity in deep networks: naive pair enumeration scales with the product of pre- and post-synaptic spike counts, and even practical online implementations amortize this cost by maintaining per-synapse eligibility traces and event-driven updates -- i.e., extra state variables and reads/writes for every active connection -- whose footprint grows with depth, fan-in/fan-out, and batch size \cite{clopath2010connectivity,pfister2006triplets,florian2007reinforcement,pfeiffer2018deep}. Moreover, purely local timing rules provide no solution to deep credit assignment, making stable supervised training in multi-layer settings difficult without additional mechanisms \cite{zenke2018superspike,pfeiffer2018deep}. 
These limitations stand in contrast to growing biological evidence that neuronal synchrony, defined as the coordinated firing of groups of neurons, plays a vital role in efficient information processing and learning.

Synchrony among neurons is widely observed across various brain regions during cognitive processes such as memory recall, sensory binding, and attention modulation \cite{engel2001dynamic, singer1999neuronal}. This synchronized activity is not accidental; it is believed to play a key role in neural coding by linking distributed neural activities and strengthening the reliability of neural communication \cite{brette2012computing, arenas2008synchronization}. Both theoretical and experimental studies indicate that neural synchrony improves response reliability, enhances plasticity thresholds, and contributes to stable attractor dynamics during learning \cite{baruchi2008emergence, kembay2025quantitative, huguenard2007thalamic}. Moreover, studies have demonstrated that precisely timed synchrony among groups of neurons can trigger long-term potentiation (LTP) even in the absence of pairwise spike causality \cite{anisimova2023spike,frey1997synaptic,markram1997regulation}. These findings challenge the sufficiency of pair-based STDP rules and suggest that collective spike patterns, rather than isolated spike timings, might be more biologically relevant for driving synaptic changes \cite{krueger1993neuronal, patel2023local}.


 We propose \textbf{S}pike-\textbf{S}ynchrony-\textbf{D}ependent \textbf{P}lasticity (SSDP), a minimal training-time local rule that turns population synchrony into a structure-aware signal. SSDP adjusts synaptic weights according to the degree of coordinated pre–post firing within a short window, using a symmetric Gaussian kernel so that near-coincident spikes still contribute to potentiation. The rule complements backpropagation instead of replacing it: the forward graph remains unchanged; updates are clipped and applied to a sparse parameter subset \emph{after} the optimizer step; and the computational overhead scales linearly with neuron counts via broadcasted outer products rather than pair enumeration. This design targets three engineering goals: (i) steadier convergence with lower loss variance, (ii) more structured representations, and (iii) robustness to spike-time jitter and event noise, all at negligible compute and memory overhead.

We systematically evaluate SSDP across diverse tasks and network architectures, showing that it enhances learning stability, reduces convergence volatility, and improves classification accuracy. We further integrate SSDP into deeper architectures, including Transformer-based spiking models, and trained deep spiking neural networks on challenging classification tasks, spanning static images, event-driven data, and datasets with high temporal structure such as SHD and SSC \cite{cramer2020heidelberg}. Together, these results establish SSDP as a biologically plausible and generalizable training signal for deep SNNs.

\begin{figure}[t]
\centering
\includegraphics[width=0.8\linewidth]{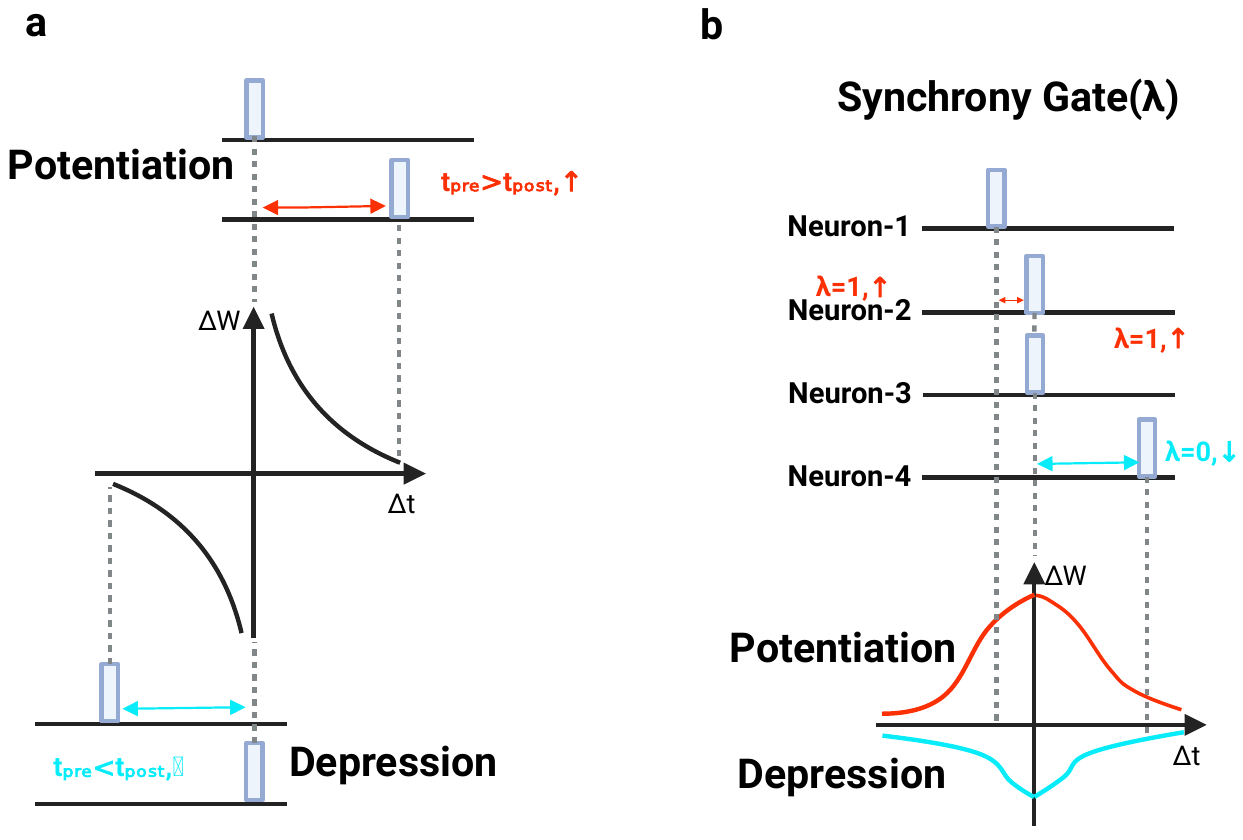}
\caption{%
\textbf{STDP versus SSDP.} 
\textbf{(a) Classical STDP.} For a synapse $i\!\to\!j$, the signed delay
$\Delta t=t_{\text{post}}-t_{\text{pre}}$ sets the \emph{sign} of the update:
pre $\!\to\!$ post ($\Delta t>0$) yields potentiation, post $\!\rightarrow$ pre
($\Delta t<0$) yields depression; the magnitude decays with $|\Delta t|$
(two-sided, order-sensitive window; schematic).
\textbf{(b) SSDP.} A binary \emph{synchrony gate} $\lambda=P_j Q_i$ (AND of post/pre spike flags) marks whether both neurons fire at least once within the same sample window. If $\lambda=1$ the pair is potentiated; if $\lambda=0$ a small depression is applied. The update magnitude is scaled by a \emph{symmetric Gaussian} $g(|\Delta t|)$ centred at $0$ using the \emph{first-spike} times, so the rule is insensitive to firing order and rewards temporal coincidence rather than causality. SSDP thus needs only one bit and one timestamp per neuron.}
\label{fig:SSDP}
\end{figure}

\section{Methods}
\label{sec:ssdp-method}

\noindent \textbf{Preliminaries.}
Consider a layer that connects
$C_{\mathrm{in}}$ presynaptic neurons to
$C_{\mathrm{out}}$ postsynaptic neurons.
Within a simulation window of length $T$
(time steps $t=0,\dots,T-1$) and for a
minibatch of size $B$, we record the binary spike flags:
\begin{equation}
  Q_{b,i}\;=\;\mathbb 1\bigl[x^{\mathrm{pre}}_{b,i}(t)>0
               \text{ for some }t\bigr],\quad
  P_{b,j}\;=\;\mathbb 1\bigl[x^{\mathrm{post}}_{b,j}(t)>0
               \text{ for some }t\bigr],
  \label{eq:spikeflags}
\end{equation}
where $x^{\mathrm{pre}}_{b,i}(t)$ and
$x^{\mathrm{post}}_{b,j}(t)$ denote the membrane outputs
of neuron $i$ and $j$ for sample
$b\in\{1,\dots,B\}$.
In addition, we store the first-spike times
\begin{equation}
  t^{\mathrm{pre}}_{b,i}\;=\;
  \min\{t\mid x^{\mathrm{pre}}_{b,i}(t)>0\},
  \qquad
  t^{\mathrm{post}}_{b,j}\;=\;
  \min\{t\mid x^{\mathrm{post}}_{b,j}(t)>0\},
  \label{eq:firstspike}
\end{equation}
and set $t^{\mathrm{pre}}_{b,i}=t^{\mathrm{post}}_{b,j}=T$ if the neuron is silent.
Using $\{0,1\}$ spike flags focuses the update on co-activation rather than raw firing rates, which makes the rule invariant to uniform rate scaling and less sensitive to burstiness.
Recording only first-spike times provides a compact temporal anchor that preserves causal ordering with $O(BC)$ memory instead of $O(BCT)$, and it still supports a coincidence kernel based on time differences.
In implementation, $Q$ and $P$ map to binary tensors, while $t^{\mathrm{pre}}$ and $t^{\mathrm{post}}$ are cached during the forward pass; silent units default to $T$, which yields large $|\Delta t|$ and therefore negligible coincidence strength in the kernel used later.

\smallskip
\noindent \textbf{Synchrony gate.}
For every pair $(j,i)$ we introduce a binary synchrony gate:
\begin{equation}
  \lambda_{b,j,i}\;=\;P_{b,j}\,Q_{b,i},
  \label{eq:lambda}
\end{equation}
which equals $1$ if and only if both neurons emit at least one spike within the same sample, and $0$ otherwise.
This gate is implemented by a broadcasted outer product of the spike-flag vectors, producing a tensor of shape $[B, C_{\mathrm{out}}, C_{\mathrm{in}}]$.
It amounts to an AND over the per-neuron temporal OR-pools $P_{b,j}$ and $Q_{b,i}$: $\lambda_{b,j,i}=1$ only when both neurons spike at least once within the window, and $0$ otherwise. The gate itself is time-agnostic once firing occurs; temporal proximity and tolerance to small jitter are governed by the coincidence kernel applied later to $|\Delta t|=|t^{\mathrm{post}}_{b,j}-t^{\mathrm{pre}}_{b,i}|$ with a learnable bandwidth $\sigma$.
The operation is fully vectorized with $O(B\,C_{\mathrm{out}}C_{\mathrm{in}})$ complexity and negligible extra memory, which matches the GPU implementation and enables stable per-minibatch averaging of the updates.

\noindent \textbf{Temporal proximity.}
The absolute delay between the first spikes is:
\begin{equation}
  \Delta t_{b,j,i}
    \;=\;
  \bigl|t^{\mathrm{post}}_{b,j}-t^{\mathrm{pre}}_{b,i}\bigr|
  \label{eq:deltat}
\end{equation}
and is converted into a smooth coincidence weight:
\begin{equation}
      g_{b,j,i}
  \;=\;
  \exp\!\Bigl(-\tfrac{\Delta t_{b,j,i}^{2}}{2\sigma^{2}}\Bigr),
\end{equation}
where $\sigma>0$ controls the temporal spread of the window.
The kernel rewards near-synchronous activity while decaying continuously with increasing $|\Delta t|$, so updates change gradually rather than flipping at a hard boundary.
Because $g$ is an even and smooth function with a finite slope at $\Delta t=0$, the rule avoids the discontinuities of rectangular windows and the cusp of Laplacian forms, which stabilizes optimization with respect to $\sigma$ and prevents sharp jumps in the update when spike times jitter.
The scale parameter $\sigma$ directly sets the coincidence tolerance: larger values widen the effective window and smaller values emphasize precise alignment; for reference, the weight drops to $1/\mathrm{e}$ at $|\Delta t|=\sigma\sqrt{2}$ and to half its peak at $|\Delta t|=\sigma\sqrt{2\ln 2}$. 
Because $g_{b,j,i}$ depends on $|\Delta t|$ only, the rule is insensitive to causal order and explicitly targets temporal coincidence instead of pre–post ordering.
Given~\eqref{eq:lambda}–\eqref{eq:deltat}, the instantaneous weight increment for sample $b$ is:
\begin{equation}
  \Delta w_{b,j,i}
  \;=\;
  \bigl(A_{+}\,\lambda_{b,j,i}
        -A_{-}\,[1-\lambda_{b,j,i}]\bigr)\,
  g_{b,j,i},
  \label{eq:dwsample}
\end{equation}
so synchronous pairs ($\lambda_{b,j,i}=1$) are reinforced in proportion to the Gaussian kernel whereas asynchronous pairs receive a mild depression.

\smallskip
\noindent \textbf{Batch aggregation and update.}
The sample-level increments are averaged over the
minibatch and applied in place:
\begin{equation}
  \Delta W_{j,i}\;=\;\frac{1}{B}\sum_{b=1}^{B}\Delta w_{b,j,i},
  \qquad
  W_{j,i}\;\leftarrow\;W_{j,i}+\Delta W_{j,i}.
  \label{eq:batchupdate}
\end{equation}
Averaging over the batch reduces gradient-like variance in the local update and makes the rule less sensitive to outlier samples with abnormal spike timing.
In practice, the implementation obtain a matrix of shape $[C_{\mathrm{out}}, C_{\mathrm{in}}]$, clips it to a bounded range to guard against rare but large coincident bursts, and then performs an in-place addition to the target weight. Updates are applied only after an initial warm-up phase so that backpropagation can establish a reasonable operating point before the synchrony prior is introduced, which improves stability without affecting the inference-time computation graph or latency.
This fully vectorized, local, and online application adds negligible training-time overhead while preserving the forward pass and optimizer state, making the procedure easy to integrate into modern SNN–Transformer training pipelines.

\noindent \textbf{Relation to gradient descent.} Steps~\eqref{eq:spikeflags}–\eqref{eq:batchupdate} are executed once per iteration \emph{after} the optimiser has performed its parameter update. The Hebbian increment is applied inside a \texttt{torch.no\_grad} context and is therefore outside the computational graph; the backpropagated gradients and the optimiser’s preconditioning or momentum buffers are computed and consumed before the SSDP tweak is added. This ordering leaves gradient flow and the optimiser’s internal state untouched within the same iteration while still modifying the weights for the next one. In practice, the additional term acts as a lightweight, local regulariser that biases synapses toward spatio-temporal co-activation without altering the forward computation or inference-time cost. 
Warm-up gating delays the application of the rule until the network has stabilized under pure backpropagation, thereby avoiding early-stage drift and preserving the convergence behaviour typically observed with SGD or Adam. Owing to its locality and reliance on batch-mean aggregation, the update integrates seamlessly into standard training loops and remains compatible with weight decay and other global regularisers.

\smallskip \subsection*{Complexity and stability} Let $B$ be the minibatch size and $C_{\mathrm{in}},C_{\mathrm{out}}$ the numbers of pre- and post-synaptic units in the plastic layer. First-spike times are denoted by $t^{\mathrm{pre}}$ and $t^{\mathrm{post}}$, and binary spike flags by $Q_{b,i}\in\{0,1\}^{B\times C_{\mathrm{in}}}$ and $P_{b,j}\in\{0,1\}^{B\times C_{\mathrm{out}}}$. Each SSDP update is built from fully vectorised broadcast operations rather than explicit event-pair enumeration. The synchrony gate arises from a single broadcasted outer product $P Q^{\top}$ with $O(B\,C_{\mathrm{out}}C_{\mathrm{in}})$ work. Pairwise time differences are formed by broadcasting $|t^{\mathrm{post}}-t^{\mathrm{pre}}|$ over indices, followed by an elementwise Gaussian kernel; both steps have the same asymptotic cost. The per-sample increments then combine the kernel with the potentiation/depression gains and are averaged across the batch to produce $\Delta W$, again with $O(B\,C_{\mathrm{out}}C_{\mathrm{in}})$ complexity. Overall, one SSDP update costs: \[ \mathcal{O}\!\bigl(B\,C_{\mathrm{out}}C_{\mathrm{in}}\bigr) \] with a small constant factor dominated by one outer product, one broadcasted subtract-and-absolute operation, one exponential, and a few fused multiplies and adds. The method stores only a single timestamp per unit and binary flags, so no factor in the average spike counts $\langle S_{\mathrm{pre}}\rangle$ or $\langle S_{\mathrm{post}}\rangle$ appears in either compute or memory; complexity remains strictly linear in neuron counts. For comparison, pair-based STDP enumerates events and scales as: \[ \mathcal{O}\!\bigl(B\,C_{\mathrm{out}}C_{\mathrm{in}}\,\langle S_{\mathrm{pre}}\rangle \langle S_{\mathrm{post}}\rangle\bigr), \] which grows with spike density and the temporal window. Stability is supported by several implementation choices that follow directly from the update structure. Batch averaging reduces variance from atypical samples and prevents large, sample-specific excursions. Clipping the aggregated increment to a bounded range guards against rare but strong coincidences and outliers in $\Delta t$. The binary gate confines potentiation to genuinely co-active pairs, while the Gaussian kernel reduces the influence of large time gaps and makes the update change smoothly as spikes jitter. Delayed activation of SSDP after a short warm-up allows backpropagation to establish a reasonable representation before the synchrony prior is introduced, which improves robustness without changing the inference graph. In the attention pathway, spatial dimensions are collapsed to channel-level indicators and first-spike times before applying the same rule, so the additional cost for extracting first-spike indices scales linearly is negligible relative to the forward pass of the network. Across each integration points, updates are in-place and local, adding minimal training-time overhead while remaining compatible with standard optimiser settings.

\noindent \textbf{Memory and traffic.} 
Beyond the model parameters, the only persistent working buffer is the batch accumulator $\Delta W \in \mathbb{R}^{C_{\mathrm{out}} \times C_{\mathrm{in}}}$, which sets the peak weight-space footprint. 
Intermediate tensors $(\lambda, \Delta t, g)$ are generated and consumed sequentially, allowing them to be streamed or tiled rather than materialised simultaneously; this keeps the peak footprint to the accumulator plus the per-neuron flags and timestamps already maintained for the forward pass. 
Each iteration requires only a single read of the spike flags and first-spike indices, followed by a single read-modify-write sweep over the target matrix; no per-synapse eligibility traces are carried across iterations. 
Bandwidth can be further reduced by exploiting data types: timestamps remain integer, flags are stored as \texttt{bool}/\texttt{uint8}, and the kernel is evaluated in the current compute dtype (AMP-compatible). 
Consequently, the SSDP step inherits mixed-precision speedups without additional casting of cached indices. 
Within the attention pathway, retaining only channelwise binary indicators and first-spike indices (rather than full feature maps) ensures that the stored activations scale as $O(TBC)$ and remain independent of $H \times W$ after the logical AND, thereby alleviating both memory pressure and traffic during the SSDP update.

\noindent \textbf{Implementation details.} Full implementation details, including data preprocessing, architectures, optimiser/schedule, SSDP update settings are provided in Supplementary (\ref{sup3}), Supplementary (\ref{sup4}) and Supplementary (\ref{sup5}).
\section{Results}
\begin{table}[!t]
\centering
\caption{Classification accuracy (Top-1) on ImageNet-1K}
\label{tab:imagenet-main}
\rowcolors{2}{white}{gray!10}
\begin{adjustbox}{max width=\linewidth}
\begin{tabular}{lcccccc}
\hline
 & \textbf{Method} & \textbf{Architecture} & \textbf{Param (M)} & \textbf{Energy (mJ)} & \textbf{TimeStep} & \textbf{Accuracy (\%)}  \\
\hline
 & Hybrid training \cite{rathi2020enabling} & ResNet-34 & 21.79 & - & 350 & 61.48 \\
 & Spiking ResNet \cite{hu2021spiking} & ResNet-50 & 25.56 & 70.934 & 350 & 72.75 \\
 & Transformer & Transformer-8-512 & 29.68 & 38.340 & 1 & 80.80 \\
 & Spikformer & Spikformer-8-384 & 16.81 & 7.734 & 4 & 70.24 \\
 & Spikformer & Spikformer-8-768 & 66.34 & 21.477 & 4 & 74.81 \\
 & Spikingformer & Spikingformer-8-512 & 29.68 & 7.46 & 4 & 74.79 \\
 & Spikingformer & Spikingformer-8-768 & 66.34 & 13.68 & 4 & 75.85 \\
 & Spike-driven Transformer & Spike-driven Transformer-8-768 & 66.34 & 6.09 & 4 & 76.32 \\
 & SpikingResformer & SpikingResformer-L & 60.38 & 8.76 & 4 & 78.77 \\
 & \textbf{Proposed} & SpikingResformer-L & \textbf{60.38} & \textbf{8.89} & 4 & \textbf{79.35$_{\pm 0.36}$} \\
\hline
\end{tabular}
\end{adjustbox}
\end{table}
\subsection{SSDP scales across Tasks, Modalities, and Temporal Complexities}
We evaluate SSDP extensively across multiple benchmark datasets and network architectures, including static image datasets (Fashion-MNIST \cite{xiao2017fashion}, CIFAR-10, CIFAR-100 \cite{krizhevsky2009learning}, ImageNet \cite{krizhevsky2012imagenet}), event-driven neuromorphic vision datasets (N-MNIST, CIFAR10-DVS \cite{li2017cifar10}), and auditory high temporal datasets (SHD, SSC \cite{cramer2020heidelberg}). This breadth is intended to test whether a synchrony-aware training signal generalizes across distinct input statistics while keeping the inference graph unchanged; we expect consistent gains when co-activation emerges. Two biologically inspired temporal kernels (exponential decay and symmetric Gaussian) are implemented to explore different synchrony-sensitivity profiles in SSDP. For detailed experimental configurations and hyperparameter settings, please refer to the Supplementary Material  (\ref{sup1}) and Supplementary Material  (\ref{sup2}).

\noindent \textbf{ImageNet-1k.}
ImageNet is a scale stress-test: large label space and high-resolution inputs require long training schedules, and compared with event-driven datasets, the activations in our spiking Transformer are relatively dense, making it a suitable benchmark for testing SSDP’s low-overhead integration. SSDP’s update is linear in neuron counts and applied after the optimizer without altering the forward graph, so we expect negligible overhead and a regularising effect from event-structure priors at scale. With bounded, Gaussian-weighted synchrony updates as a light-weight regulariser, accuracy should improve over the backprop-only baseline while training speed remain unchanged. Table~\ref{tab:imagenet-main} presents ImageNet-1k results with the SpikingResformer-L backbone ($T{=}4$). Under matched settings, DA-SSDP improves the baseline and attains competitive performance relative to recent spiking Transformers of comparable size, consistent with the expectation that a synchrony-aware training signal scales to large-vocabulary recognition without modifying the forward computation.

\noindent \textbf{CIFAR-10/100 and CIFAR10-DVS.} We use two static image benchmarks and an event-stream dataset to test whether SSDP’s benefits persist across input sparsity regimes and temporal structure. On CIFAR-10/100, spatial regularities tend to elicit repeatable channel co-activation, so synchrony-weighted updates should reinforce useful representations. On CIFAR10-DVS, events are produced by scan-converting static images; the resulting streams are sparse and bursty with weak class-aligned timing across a mini-batch, so the synchrony gate \(\lambda = P\,Q^{\top}\) rarely activates at zero lag and the Gaussian factor \(g=\exp\!\bigl(-\Delta t^{2}/(2\sigma^{2})\bigr)\) is typically small -- hence gains may be limited. Table~\ref{tab:smalldatasets} reports results using CIFAR variants of SpikingResformer: SSDP consistently matches or exceeds the backprop-only baseline on CIFAR-10/100 under identical hyperparameters, while improvements on CIFAR10-DVS are modest, consistent with the weaker synchrony signal.

\noindent \textbf{Temporal auditory benchmarks.}
SHD and SSC carry explicit time structure (onset patterns, inter-spike intervals) that repeatedly co-activate subsets of channels; under such inputs, a synchrony-weighted update should reinforce task-relevant assemblies. Table~\ref{tab:SRNN-results} compares SRNN-family models on SHD and SSC. Adding SSDP to the DHSRNN backbone improves Top-1 on both datasets while keeping the non-convolutional structure unchanged, indicating that synchrony-guided local updates transfer to temporal classification and are not tied to convolutional feature extractors.
\begin{wraptable}{r}{0.5\textwidth} 
\vspace{-\baselineskip}             
\centering
\caption{Classification accuracy (Top-1) based on SNNs.}
\label{tab:SRNN-results}

\rowcolors{2}{white}{gray!10}
\resizebox{\linewidth}{!}{          
\begin{tabular}{llcc}
\hline
\textbf{Dataset} & \textbf{Learning rule} & \textbf{Conv. structure} & \textbf{Accuracy (\%)} \\
\hline
\textbf{SHD} & BrainScaleS-2 \cite{bib46} & \textcolor{red}{\ding{55}} & 81.6 \\
& SRNN \cite{bib49} & \textcolor{red}{\ding{55}} & 82.7 \\
& SCNN \cite{bib50} & \textcolor{green}{\ding{51}} & 84.8 \\
& SRNN \cite{bib48} & \textcolor{red}{\ding{55}} & 90.4 \\
& LSTM \cite{bib45} & \textcolor{red}{\ding{55}} & 89.2 \\
& DHSRNN \cite{exp1} & \textcolor{red}{\ding{55}} & 88.1 \\
& \textbf{DHSRNN+SSDP} & \textcolor{red}{\ding{55}} & \textbf{89.1$_{\pm 0.21}$} \\
\hline
\textbf{SSC} 
& SRNN & \textcolor{red}{\ding{55}} & 74.2 \\
& LSTM & \textcolor{red}{\ding{55}} & 73.1 \\
& DHSNN & \textcolor{red}{\ding{55}} & 82.46 \\
& \textbf{DHSNN+SSDP} & \textcolor{red}{\ding{55}} & \textbf{82.86$_{\pm 0.26}$} \\
\hline
\end{tabular}
}
\end{wraptable}
Across large-scale static vision and temporal auditory tasks, SSDP provides consistent accuracy gains with no inference-time structural cost and minimal implementation burden. To attribute changes solely to the training signal, we hold architecture and schedules fixed across conditions. Unless otherwise stated, results are reported as Top-1 accuracy over $N{=}5$ independent seeds. For each seed, the checkpoint with the highest validation accuracy is evaluated on the test set. All comparisons use identical training schedules (epochs, batch size, optimizer, time steps) and unchanged architectures and parameter counts. SSDP operates only during training and introduces no inference-time structural overhead. Additional single–hidden-layer SNN experiments on Fashion-MNIST, CIFAR-10, and N-MNIST are provided in the Supplementary Material (\ref{sup3}).

\begin{figure}[!t]
    \centering
    \includegraphics[width=0.9\linewidth]{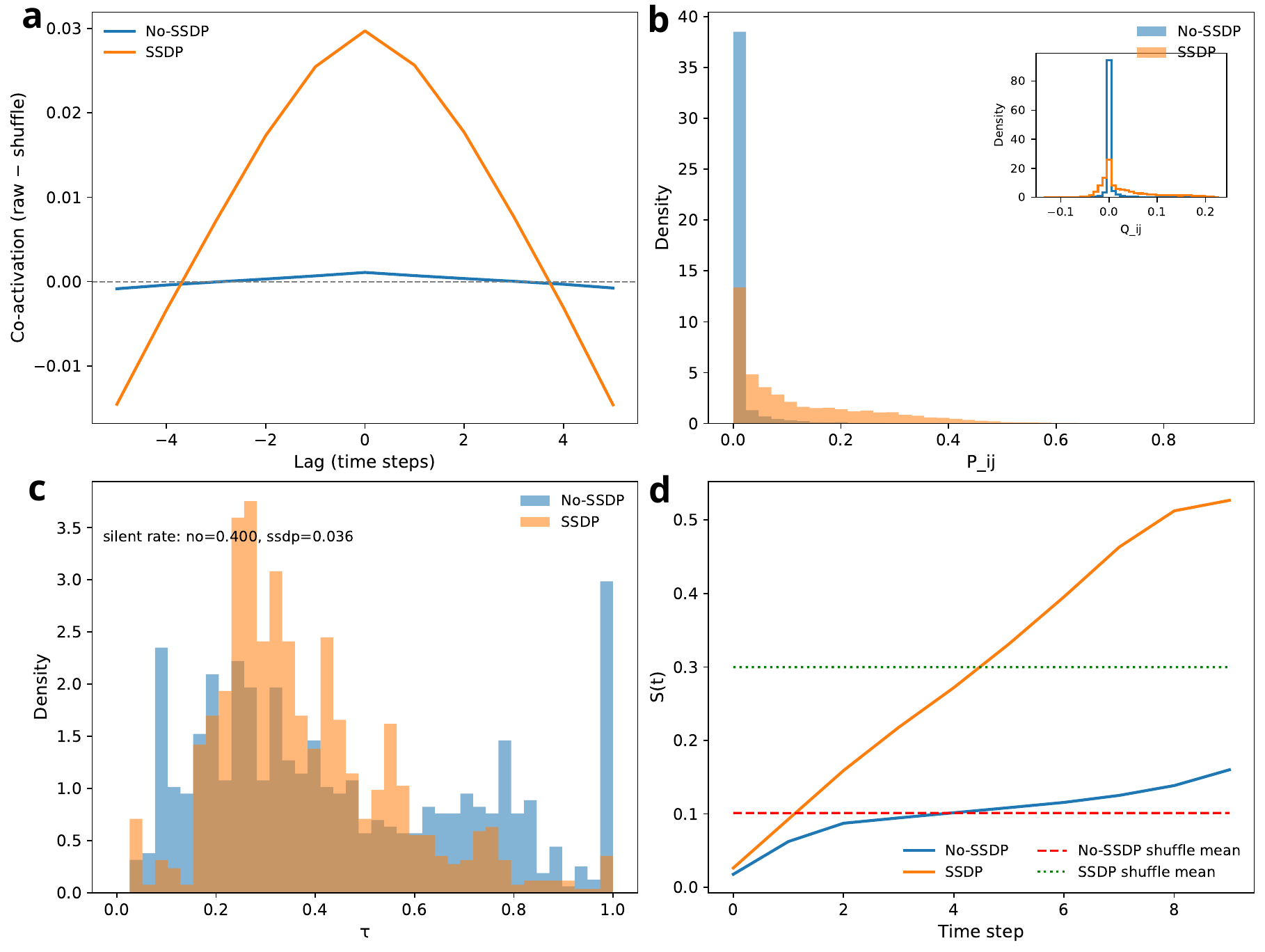}
    \caption{\textbf{(a)} Excess pairwise co-activation (raw-shuffle) versus lag; SSDP shows a sharp symmetric peak at zero lag, whereas the baseline stays near zero, indicating near-simultaneous co-activation beyond rate effects. 
    \textbf{(b)} Distributions of zero-lag co-activation $P_{ij}=\Pr[s_i(t)=1\wedge s_j(t)=1]$ across the same randomly sampled pairs (main) and the rate-corrected statistic $Q_{ij}=P_{ij}-p_i p_j$ with $p_i=\Pr[s_i(t)=1]$ (inset); both shift to the right under SSDP, showing the increase is not explained by marginal firing rates. 
    \textbf{(c)} Temporal selectivity for non-silent units, $\tau_j=1-H_j/\log T$ with $H_j$ the entropy of the rectified, time-normalized within-unit $z$-score; the silent-unit fraction drops from $0.400$ to \textbf{$0.036$}, and mass concentrates at moderate $\tau$, indicating broader recruitment with non-impulsive tuning. 
    \textbf{(d)} Population activity $S(t)=\tfrac{1}{BN}\sum_{b,j}s_{b,t,j}$ for each condition together with its own time-shuffled expectation; SSDP lies consistently above its shuffle band and diverges most strongly late, indicating stronger time-localised recruitment. All statistics are computed from recorded spike tensors. Shuffle surrogates use independent circular shifts per unit for (a–b) and independent time permutations per unit for (d), preserving marginal rates while destroying cross-unit timing.}
    \label{fig:ssdp_synchrony}
    
\end{figure}
\begin{table}
\centering
\caption{Comparision on CIFAR-10, CIFAR-100, CIFAR10-DVS.}
\label{tab:smalldatasets}

\rowcolors{2}{white}{gray!10}
\begin{adjustbox}{max width=\linewidth}
\begin{tabular}{lccccc}
\hline
\textbf{Dataset} & \textbf{Method} & \textbf{Architecture} & \textbf{Param (M)} &  \textbf{TimeStep} & \textbf{Accuracy (\%)}  \\
\hline
\textbf{CIFAR-10}
 & Spikformer \cite{zhou2022spikformer} & Spikformer-4-384 & 9.32  & 4 & 95.51 \\
 & Spikingformer \cite{zhou2023spikingformer} & Spikingformer & -  & 4 & 95.61 \\
 & Spike-driven Transformer \cite{yao2024spike} & Spike-driven Transformer &  - & 4 & 95.6 \\
 & Transformer \cite{dosovitskiy2020image} & Transformer-4-384 & 9.32  & 1 & 96.73 \\
 & \textbf{Proposed} & SpikingResformer-Cifar & \textbf{10.83}  & 4 & \textbf{96.24$_{\pm 0.29}$} \\
\hline
\textbf{CIFAR-100}
 & Spikformer & Spikformer-4-384 & 9.32  & 4 & 77.86 \\
 & Spikingformer & Spikingformer & -  & 4 & 79.09 \\
 & Spike-driven Transformer & Spike-driven Transformer & -  & 4 & 78.4 \\
 & Transformer & Transformer-4-384 & 9.32 & 1 & 81.02 \\
 & SpikingResformer \cite{shi2024spikingresformer} & SpikingResformer-Cifar & 10.83 &  4 & 78.73 \\
 & \textbf{Proposed} & SpikingResformer-Cifar & \textbf{10.83} & 4 & \textbf{79.48$_{\pm 0.27}$} \\
\hline
\textbf{CIFAR10-DVS}
 & Spikformer & Spikformer-4-384 & 9.32  & 16 & 80.6 \\
 & Spikingformer & Spikingformer & -  & 16 & 81.3 \\
 & Spike-driven Transformer & Spike-driven Transformer & -  & 16 & 80.0 \\
 & Transformer & Transformer-4-384 & 9.32  & 1 & 81.02 \\
 & \textbf{Proposed} & SpikingResformer-Cifar & \textbf{17.31} & 10 & \textbf{84.5$_{\pm 0.3}$} \\
\hline
\end{tabular}
\end{adjustbox}
\end{table}
\subsection{SSDP restructures population timing under rate control}
Having established accuracy gains, we next examine how SSDP changes network dynamics in ways that could plausibly support those gains. The working hypothesis is that SSDP acts as a lightweight event-structure regulariser: it should increase near-simultaneous co-activation and organise population recruitment in time. To isolate coordination beyond rate, each model is compared to a time-shuffled surrogate that preserves marginal rates while disrupting cross-unit timing. Fig.~\ref{fig:ssdp_synchrony} aggregates four complementary readouts: Fig.~2~(a–b) provide rate-controlled pairwise evidence; Fig.~2~(c–d) describe unit recruitment and its population-level manifestation.

\noindent\textbf{Fig.~2~(a).}
For each lag $\ell\!\in\![-L,L]$, we average over random neuron pairs the probability that one unit spikes at $t$ while the other spikes at $t{+}\ell$, and subtract the shuffled surrogate. Under SSDP a sharp, symmetric zero-lag peak emerges, whereas the baseline remains near zero. Because the shuffle controls for marginal rates, this peak indicates same-time synchrony activities beyond rate effects.

\noindent\textbf{Fig.~2~(b).}
We report the distribution of $P_{ij}{=}\Pr[s_i(t){=}1\wedge s_j(t){=}1]$ across the same randomly sampled pairs (main), together with the rate-corrected statistic $Q_{ij}{=}P_{ij}{-}p_ip_j$ where $p_i{=}\Pr[s_i(t){=}1]$ (inset).
Both distributions shift right under SSDP, indicating that the increase in zero-lag coincidence is not explained by the product of marginal firing rates and is consistent with Fig. 2 (a). This rightward shift in both $P_{ij}$ and the rate-corrected $Q_{ij}$ matches the expectation that coincidence increases beyond what marginal rates predict.

\noindent\textbf{Fig.~2~(c).}
For each non-silent unit we compute $m_j(t)$ (spike probability across batches), standardize it over $t$, truncate to the positive part, renormalize to $p_j(t)$, and set $H_j{=}{-}\sum_t p_j(t)\log p_j(t)$ and $\tau_j{=}1{-}H_j/\log T$ (larger $\tau_j$ indicates narrower, phase-focused activity). We expect broader recruitment with moderate temporal selectivity rather than sparse extremes. Under SSDP the silent-unit fraction drops from $0.400$ to \textbf{$0.036$}, and the non-silent population concentrates at moderate $\tau$, indicating broader recruitment with well-defined but reasonable temporal tuning. 

\noindent\textbf{Fig.~2~(d).}
Let $S(t){=}\tfrac{1}{BN}\sum_{b,j}s_{b,t,j}$ denote the fraction of active units per step.
For each condition we also plot the expectation from its own time-permutation surrogate.
Both conditions lie above their shuffle bands, and SSDP diverges most strongly at later steps, consistent with stronger time-localised population recruitment. Note that $S(t)$ reflects activity level rather than a pairwise synchrony metric; Fig.~2~(a–b) constitute the synchrony evidence, while Fig.~2~(d) shows its population-scale expression.

\noindent\textbf{Link to performance.} 
Taken together, these effects are consistent with SSDP organising spikes into temporally coordinated population events -- an arrangement that can reduce training-loss variance and support generalisation, as observed in our convergence analysis.

\subsection{Representation structure and temporal precision}

\begin{figure}
    \centering
    \begin{minipage}[t]{0.48\linewidth}
        \centering
        \includegraphics[width=\linewidth]{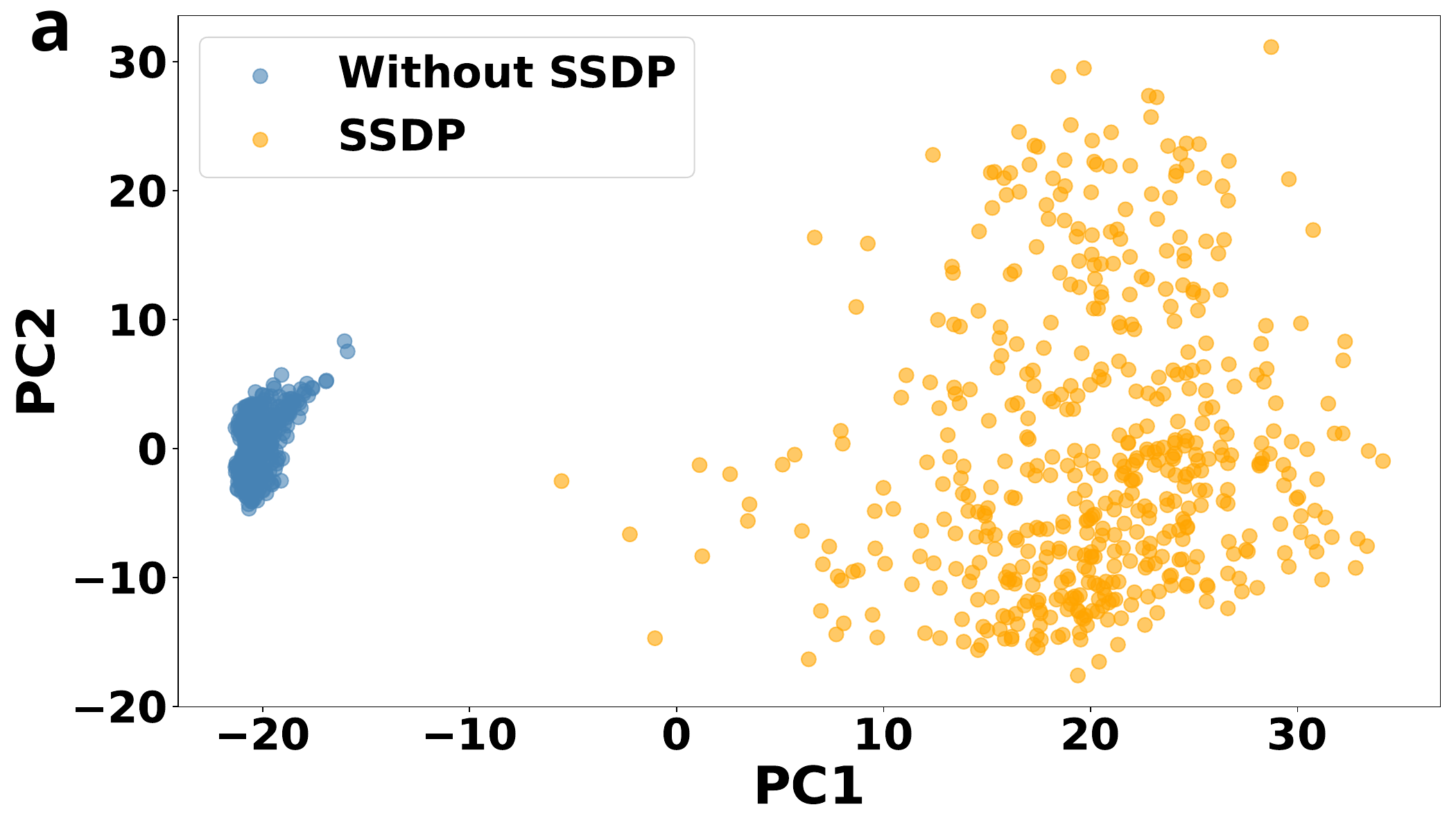}
    \end{minipage}
    \hfill
    \begin{minipage}[t]{0.48\linewidth}
        \centering
        \includegraphics[width=\linewidth]{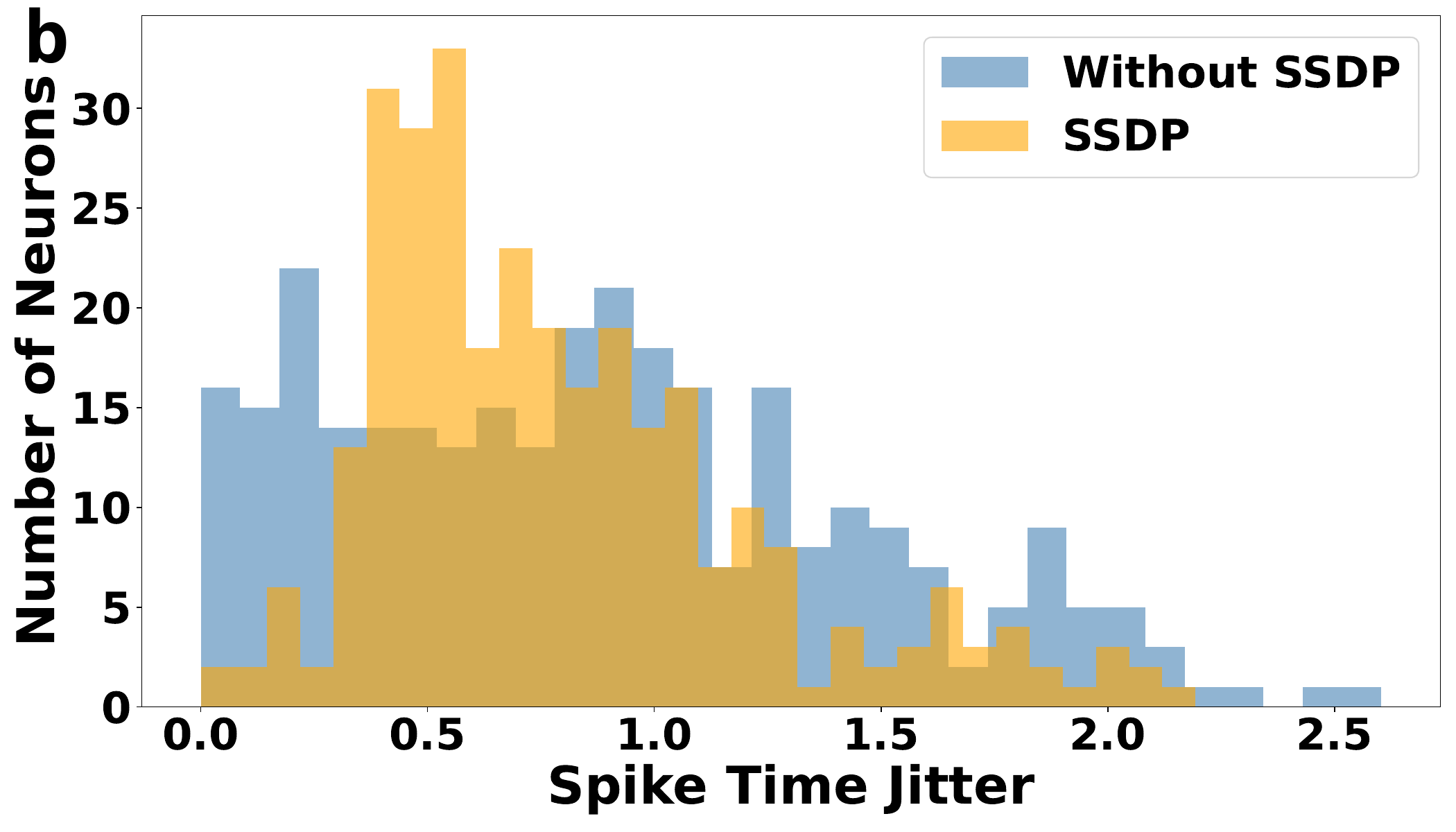}
    \end{minipage}

    \caption{\textbf{(a)}: PCA of the same feature layer under matched training shows that SSDP (orange) yields a broader distribution with a pronounced principal direction compared to the baseline (blue).
    \textbf{(b)}: 
    Compared with the baseline, SSDP also reduces dispersion (median $0.699$ vs.\ $0.846$, IQR $0.523$ vs.\ $0.853$) and shrinks the high-jitter tail ($>\!1.5$: $0.083$ vs.\ $0.160$; $>\!1.0$: $0.263$ vs.\ $0.380$), while avoiding near-zero jitter saturation ($\approx 0$: $0.003$ vs.\ $0.040$). 
    A two-sample KS test detects a distributional shift ($p{=}9{\times}10^{-4}$); Mann-Whitney on medians is not significant ($p{=}0.185$), consistent with changes occurring mainly in the tails and shape rather than a large central shift.}
    \label{fig:combined}
\end{figure}

\paragraph{Linear embedding evidence.}
As a linear, model-agnostic probe, PCA is used to test the hypothesis that SSDP reorganises population activity from diffuse, many-axis variability into shared, low-rank structure. If SSDP reinforces co-activation, the covariance should concentrate along a few common directions, yielding higher total variance carried by leading components and a lower participation ratio (rather than a uniform spread across weak axes). This pattern is desirable because aligned dominant modes support more stable downstream optimisation and cleaner class structure. The outcomes in Fig.~\ref{fig:combined}(a) -- increased total variance together with a reduced participation ratio -- are consistent with this expectation and quantify the representation change induced by SSDP.
Quantitatively, in the standardised feature space the total variance increases markedly under SSDP ($811.8$ vs.\ $374.5$, $\approx 2.17\times$ in this run), while the participation ratio (effective dimensionality) decreases ($17.19$ vs.\ $23.33$). Taken together, this indicates that SSDP does not spread variability uniformly across many weak axes; instead, it concentrates more of the variability onto a smaller set of shared principal directions, yielding a more anisotropic, low-rank structure. Consistent with this, PC1 and PC2 account for roughly $0.43$ and $0.06$ of the variance ($\sim 49\%$ total), so the visible widening aligns with a spectrum where a few leading components carry most of the energy, and the remaining structure lies in higher dimensions. Another visual tests are included in Supplementary  (\ref{sup7})


\paragraph{Temporal precision (jitter).}
Temporal precision is a key property of reliable population codes. To quantify it, per-neuron jitter is computed as the across-repeat standard deviation of mean spike times (300 randomly sampled neurons per condition, fixed seed). Relative to the baseline, SSDP yields lower and tighter jitter (median $0.699$ vs.\ $0.846$; IQR $0.523$ vs.\ $0.853$; std $0.420$ vs.\ $0.575$) and markedly suppresses the high-jitter tail ($>\!1.5$: $0.083$ vs.\ $0.160$; $>\!1.0$: $0.263$ vs.\ $0.380$). At the same time, the fraction of near-zero jitter decreases ($0.003$ vs.\ $0.040$), indicating that units remain responsive rather than becoming rigid. A two-sample KS test detects a distributional change ($p{=}9{\times}10^{-4}$), whereas the Mann–Whitney test on medians is not significant ($p{=}0.185$) and Cliff’s $\delta$ is small ($-0.063$), consistent with the improvement arising from tail suppression and overall tightening rather than a large median shift (Fig.~\ref{fig:combined}(b)). Taken together with the PCA evidence (higher total variance and lower participation ratio; Fig.~\ref{fig:combined}(a)), the jitter results indicate that SSDP produces population codes that are structurally organised (variance concentrated along a few shared principal directions) and temporally stable (reduced dispersion and tails in spike timing), aligning with the goal of improving representation structure and temporal precision without altering the forward computation.
\begin{figure}[!b]
\centering
\includegraphics[width=0.85\linewidth]{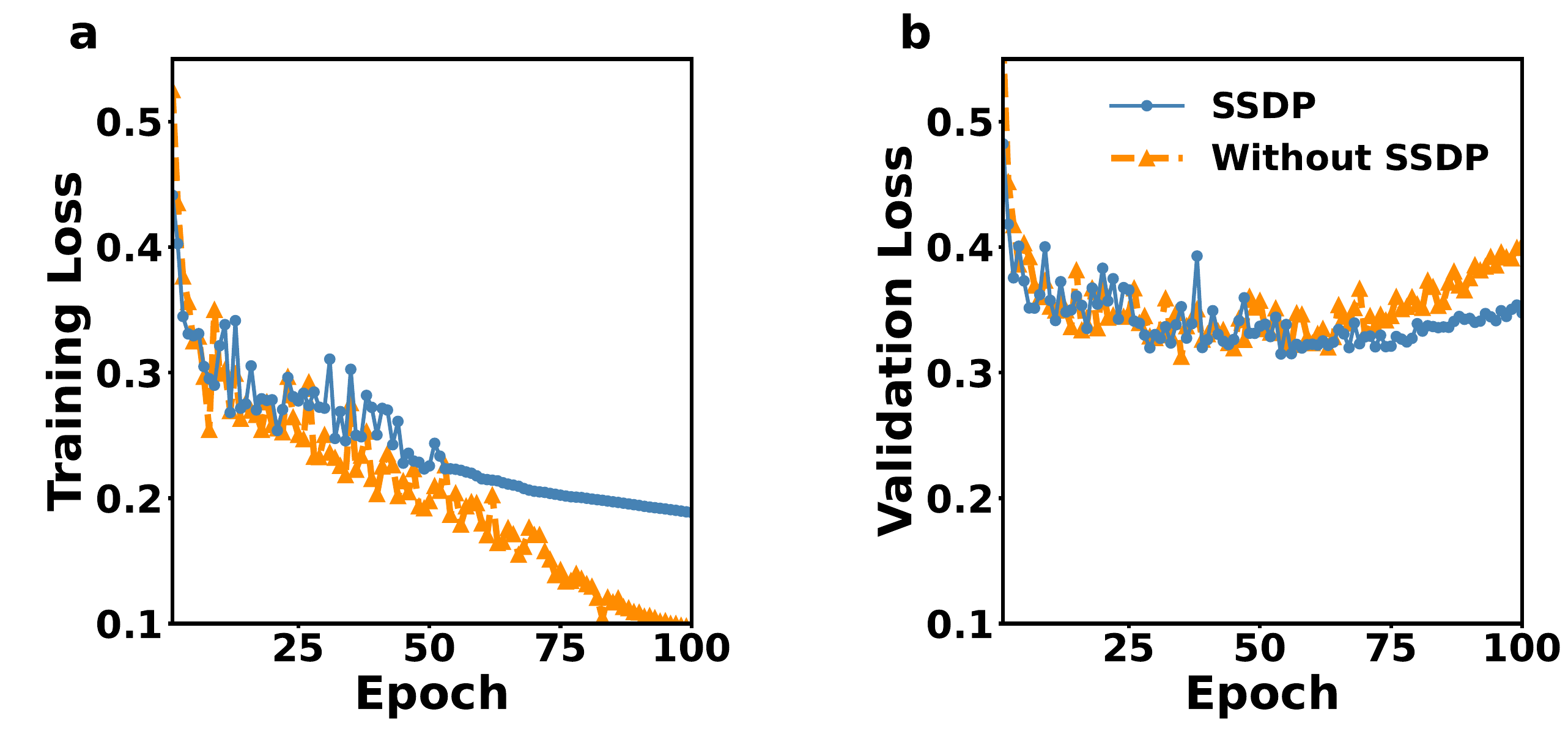}
\caption{\textbf{(a)} Training loss On Fashion-MNIST; \textbf{(b)} validation loss on Fashion-MNIST.
The SSDP model (blue) exhibits a mid-training transition to a smoother, low-variance regime, while the baseline (orange) remains more oscillatory and develops a larger train–val gap. This behaviour is consistent with SSDP’s symmetric, synchrony-based post-update, which regularises training using spike-event structure.
}
\label{fig:spike-train3}
\end{figure}

\subsection{Convergence stabilisation with SSDP}
Fig.~\ref{fig:spike-train3} summarises optimisation behaviour on Fashion-MNIST. In Fig.4 (a), both models reduce training loss initially; the baseline (orange) keeps decreasing but remains oscillatory, whereas the SSDP model (blue) enters a smoother, low-variance regime mid-training and converges steadily thereafter. In Fig.~4~(b), the SSDP model sustains a lower validation loss, while the baseline develops a widening train–validation gap -- clear evidence of overfitting. Taken together with Fig.~\ref{fig:ssdp_synchrony}, the curves indicate that SSDP reduces loss variance and stabilises convergence, acting as a lightweight regulariser that biases learning toward repeatable, task-relevant spike-event structure.

These observations align with our objectives: (i) steadier convergence, reflected in lower loss variance and a reduced train–validation gap; (ii) more structured representations, supported by rate-controlled synchrony (Fig.~\ref{fig:ssdp_synchrony}) and the PCA/jitter analyses (Fig.~\ref{fig:combined}); and (iii) greater tolerance to spike-time perturbations, implied by the tightened jitter distribution. The stabilising effect is consistent with the neuroscience view that population synchrony enhances communication and reduces representational noise \cite{fries2005mechanism,averbeck2006neural,engel2001dynamic,brette2012computing}.

\section{Discussion}

In this paper, we introduce a synchrony-sensitive, training-time local rule termed SSDP for spiking networks and integrate it into both shallow models and transformer-style SNNs; to the best of our knowledge, this is the first deployment of a biologically inspired plasticity rule within an SNN-Transformer under matched modern training budgets. Across benchmarks, SSDP improves top-1 accuracy, exhibits a mid-training transition to a lower-variance loss regime, yields broader yet organised feature geometry, and tightens spike-time reliability by shrinking high-jitter tails without enforcing rigidity. Mechanistically, SSDP operates on population co-activation within a short coincidence window: near-synchronous pre–post activity receives potentiation and non-coactive pairs receive weak depression. The update is formed by broadcasted outer products, adds linear-time memory/compute only during training.

\textbf{SSDP imposes a low-rank, synchrony-aligned prior on representations.} The rate-controlled analyses show a symmetric zero-lag excess that is not explained by marginal rates, and PCA-based probes indicate a reallocation of variance toward a few shared directions rather than a diffuse spread. This suggests that SSDP concentrates variability along population modes that recur across samples, yielding more organised feature subspaces.

\textbf{SSDP functions as a data-adaptive regulariser that scales with evidence of synchrony.} When class-aligned coincidences are abundant (e.g., static vision benchmarks), SSDP strengthens the corresponding pathways and improves accuracy; when synchrony carries little task signal (e.g., pseudo-temporal event streams), the effect remains small, leaving performance essentially neutral. This behaviour is consistent with an event-structure prior that activates only where the data warrant it.

\textbf{SSDP stabilises optimisation by aligning weight changes with repeatable spike motifs.} Because updates are restricted to co-active connections and clipped in magnitude, the per-batch parameter change is sparser and more consistent with stable motifs rather than step-to-step gradient noise, explaining the observed reduction in loss variance and train–validation gap.

\textbf{Future direction -- toward neuromorphic deployment and new AI chip.}
An important direction for future development of this innovative idea is the integration of the proposed SSDP learning rule into a custom Application-Specific Integrated Circuit (ASIC) hardware. Since SSDP operates solely during training and does not interfere with inference, a practical strategy is to apply SSDP offline during the training phase and then deploy the resulting fixed weights onto inference-only neuromorphic chips. This approach avoids any runtime hardware overhead while retaining the stability and performance improvements shown in this work. For scenarios that demand adaptability, a lightweight on-chip SSDP engine could also be introduced. Such an implementation would require only minimal per-neuron state, for example, spike presence flags and first-spike timestamps together with a small accumulator buffer for weight updates, eliminating the need for costly eligibility traces or exhaustive spike pair tracking. From a hardware perspective, realizing SSDP in silicon carries clear implications for power, memory, and scalability. The additional overhead grows linearly with neuron counts, as the update relies on a broadcasted outer product and a Gaussian kernel that can be efficiently approximated with a lookup table or simple piecewise function. Employing tiling strategies and clipped weight updates further helps to bound and manage memory usage. Importantly, because SSDP leaves the forward computation untouched, inference energy and throughput remain equivalent to conventional spiking accelerators, making it a cost-effective enhancement. Consequently, the framework presented here can be adopted in neuromorphic ASICs either as an offline training tool to improve models before deployment or as a compact on-chip mechanism enabling continual learning in dynamic edge environments.

In practice, SSDP functions as a lightweight event-structure regulariser that biases learning toward repeatable, task-relevant spike motifs while preserving convergence behaviour. Overall, injecting an event-structure signal at training time yields more organised spatial codes and more reliable spike timing under realistic budgets, and remains compatible with neuromorphic deployment due to locality and zero-cost inference. SSDP thus offers a practical mechanism for improving SNN learning while providing a testable computational hypothesis for population-level plasticity.

\section*{Data and Code Availability}

All datasets used in this study are publicly available benchmark datasets. No proprietary or custom datasets were used or created in this work. The code is available at: \url{https://github.com/NeuroSyd/SSDP}. 

\section*{Acknowledgement}
The authors acknowledge partial support from the Australian Research Council under Project DP230100019.

\small
\bibliographystyle{plain} 
\bibliography{references}

\begin{thebibliography}{10}

\bibitem{anisimova2023spike}
Margarita Anisimova, Bas van Bommel, Rui Wang, Marina Mikhaylova, J{\"o}rn~Simon Wiegert, Thomas~G Oertner, and Christine~E Gee.
\newblock Spike-timing-dependent plasticity rewards synchrony rather than causality.
\newblock {\em Cerebral Cortex}, 33(1):23--34, 2023.

\bibitem{apolinario2023s}
Marco Paul~E Apolinario and Kaushik Roy.
\newblock S-tllr: Stdp-inspired temporal local learning rule for spiking neural networks.
\newblock {\em ArXiv preprint arXiv:2306.15220}, 2023.

\bibitem{arenas2008synchronization}
Alex Arenas, Albert D{\'\i}az-Guilera, Jurgen Kurths, Yamir Moreno, and Changsong Zhou.
\newblock Synchronization in complex networks.
\newblock {\em Physics Reports}, 469(3):93--153, 2008.

\bibitem{averbeck2006neural}
Bruno~B Averbeck, Peter~E Latham, and Alexandre Pouget.
\newblock Neural correlations, population coding and computation.
\newblock {\em Nature Reviews Neuroscience}, 7(5):358--366, 2006.

\bibitem{baruchi2008emergence}
Itay Baruchi, Vladislav Volman, Nadav Raichman, Mark Shein, and Eshel Ben-Jacob.
\newblock The emergence and properties of mutual synchronization in in vitro coupled cortical networks.
\newblock {\em European Journal of Neuroscience}, 28(9):1825--1835, 2008.

\bibitem{bi1998synaptic}
Guo-qiang Bi and Mu-ming Poo.
\newblock Synaptic modifications in cultured hippocampal neurons: dependence on spike timing, synaptic strength, and postsynaptic cell type.
\newblock {\em Journal of Neuroscience}, 18(24):10464--10472, 1998.

\bibitem{bottou2010large}
L{\'e}on Bottou.
\newblock Large-scale machine learning with stochastic gradient descent.
\newblock In {\em Proceedings of COMPSTAT'2010: 19th International Conference on Computational StatisticsParis France, August 22-27, 2010 Keynote, Invited and Contributed Papers}, pages 177--186. Springer, 2010.

\bibitem{brette2012computing}
Romain Brette.
\newblock Computing with neural synchrony.
\newblock {\em PLoS Computational Biology}, 8(6):e1002561, 2012.

\bibitem{buzsaki2010neural}
Gy{\"o}rgy Buzs{\'a}ki.
\newblock Neural syntax: cell assemblies, synapsembles, and readers.
\newblock {\em Neuron}, 68(3):362--385, 2010.

\bibitem{clopath2010connectivity}
Claudia Clopath, Lars B{\"u}sing, Eleni Vasilaki, and Wulfram Gerstner.
\newblock Connectivity reflects coding: a model of voltage-based stdp with homeostasis.
\newblock {\em Nature Neuroscience}, 13(3):344--352, 2010.

\bibitem{cohen2016skimming}
Gregory~K Cohen, Garrick Orchard, Sio-Hoi Leng, Jonathan Tapson, Ryad~B Benosman, and Andr{\'e} Van~Schaik.
\newblock Skimming digits: neuromorphic classification of spike-encoded images.
\newblock {\em Frontiers in Neuroscience}, 10:184, 2016.

\bibitem{bib46}
Benjamin Cramer, Sebastian Billaudelle, Simeon Kanya, Aron Leibfried, Andreas Gr{\"u}bl, Vitali Karasenko, Christian Pehle, Korbinian Schreiber, Yannik Stradmann, Johannes Weis, et~al.
\newblock Surrogate gradients for analog neuromorphic computing.
\newblock {\em Proceedings of the National Academy of Sciences}, 119(4):e2109194119, 2022.

\bibitem{cramer2020heidelberg}
Benjamin Cramer, Yannik Stradmann, Johannes Schemmel, and Friedemann Zenke.
\newblock The heidelberg spiking data sets for the systematic evaluation of spiking neural networks.
\newblock {\em IEEE Transactions on Neural Networks and Learning Systems}, 33(7):2744--2757, 2020.

\bibitem{bib45}
Benjamin Cramer, Yannik Stradmann, Johannes Schemmel, and Friedemann Zenke.
\newblock The heidelberg spiking data sets for the systematic evaluation of spiking neural networks.
\newblock {\em IEEE Transactions on Neural Networks and Learning Systems}, 33(7):2744--2757, 2020.

\bibitem{table1}
Yiting Dong, Dongcheng Zhao, Yang Li, and Yi~Zeng.
\newblock An unsupervised stdp-based spiking neural network inspired by biologically plausible learning rules and connections.
\newblock {\em Neural Networks}, 165:799--808, 2023.

\bibitem{dosovitskiy2020image}
Alexey Dosovitskiy.
\newblock An image is worth 16x16 words: Transformers for image recognition at scale.
\newblock {\em ArXiv preprint arXiv:2010.11929}, 2020.

\bibitem{engel2001dynamic}
Andreas~K Engel, Pascal Fries, and Wolf Singer.
\newblock Dynamic predictions: oscillations and synchrony in top--down processing.
\newblock {\em Nature Reviews Neuroscience}, 2(10):704--716, 2001.

\bibitem{eshraghian2023training}
Jason~K Eshraghian, Max Ward, Emre~O Neftci, Xinxin Wang, Gregor Lenz, Girish Dwivedi, Mohammed Bennamoun, Doo~Seok Jeong, and Wei~D Lu.
\newblock Training spiking neural networks using lessons from deep learning.
\newblock {\em Proceedings of the IEEE}, 111(9):1016--1054, 2023.

\bibitem{fang2023spikingjelly}
Wei Fang, Yanqi Chen, Jianhao Ding, Zhaofei Yu, Timoth{\'e}e Masquelier, Ding Chen, Liwei Huang, Huihui Zhou, Guoqi Li, and Yonghong Tian.
\newblock Spikingjelly: An open-source machine learning infrastructure platform for spike-based intelligence.
\newblock {\em Science Advances}, 9(40):eadi1480, 2023.

\bibitem{fang2021deep}
Wei Fang, Zhaofei Yu, Yanqi Chen, Tiejun Huang, Timoth{\'e}e Masquelier, and Yonghong Tian.
\newblock Deep residual learning in spiking neural networks.
\newblock {\em Advances in Neural Information Processing Systems}, 34:21056--21069, 2021.

\bibitem{frey1997synaptic}
Uwe Frey and Richard~GM Morris.
\newblock Synaptic tagging and long-term potentiation.
\newblock {\em Nature}, 385(6616):533--536, 1997.

\bibitem{fries2005mechanism}
Pascal Fries.
\newblock A mechanism for cognitive dynamics: neuronal communication through neuronal coherence.
\newblock {\em Trends in Cognitive Sciences}, 9(10):474--480, 2005.

\bibitem{goupy2024neuronal}
Gaspard Goupy, Pierre Tirilly, and Ioan~Marius Bilasco.
\newblock Neuronal competition groups with supervised stdp for spike-based classification.
\newblock {\em arXiv preprint arXiv:2410.17066}, 2024.

\bibitem{goupy2024paired}
Gaspard Goupy, Pierre Tirilly, and Ioan~Marius Bilasco.
\newblock Paired competing neurons improving stdp supervised local learning in spiking neural networks.
\newblock {\em Frontiers in Neuroscience}, 18:1401690, 2024.

\bibitem{gutig2006tempotron}
Robert G{\"u}tig and Haim Sompolinsky.
\newblock The tempotron: a neuron that learns spike timing--based decisions.
\newblock {\em Nature Neuroscience}, 9(3):420--428, 2006.

\bibitem{han2020rmp}
Bing Han, Gopalakrishnan Srinivasan, and Kaushik Roy.
\newblock Rmp-snn: Residual membrane potential neuron for enabling deeper high-accuracy and low-latency spiking neural network.
\newblock In {\em Proceedings of the IEEE/CVF Conference on Computer Vision and Pattern Recognition}, pages 13558--13567, 2020.

\bibitem{han2015learning}
Song Han, Jeff Pool, John Tran, and William Dally.
\newblock Learning both weights and connections for efficient neural network.
\newblock {\em Advances in Neural Information Processing Systems}, 28, 2015.

\bibitem{hao2020biologically}
Yunzhe Hao, Xuhui Huang, Meng Dong, and Bo~Xu.
\newblock A biologically plausible supervised learning method for spiking neural networks using the symmetric stdp rule.
\newblock {\em Neural Networks}, 121:387--395, 2020.

\bibitem{hebblearning}
Donald~Olding Hebb.
\newblock {\em The organization of behavior: {A} neuropsychological theory}.
\newblock Psychology press, 2005.

\bibitem{hebb2005organization}
Donald~Olding Hebb.
\newblock {\em The organization of behavior: A neuropsychological theory}.
\newblock Psychology Press, 2005.

\bibitem{hu2021spiking}
Yangfan Hu, Huajin Tang, and Gang Pan.
\newblock Spiking deep residual networks.
\newblock {\em IEEE Transactions on Neural Networks and Learning Systems}, 34(8):5200--5205, 2021.

\bibitem{huguenard2007thalamic}
John~R Huguenard and David~A McCormick.
\newblock Thalamic synchrony and dynamic regulation of global forebrain oscillations.
\newblock {\em Trends in Neurosciences}, 30(7):350--356, 2007.

\bibitem{indiveri2015memory}
Giacomo Indiveri and Shih-Chii Liu.
\newblock Memory and information processing in neuromorphic systems.
\newblock {\em Proceedings of the IEEE}, 103(8):1379--1397, 2015.

\bibitem{jiang2023adaptive}
Tingting Jiang, Qi~Xu, Xuming Ran, Jiangrong Shen, Pan Lv, Qiang Zhang, and Gang Pan.
\newblock Adaptive deep spiking neural network with global-local learning via balanced excitatory and inhibitory mechanism.
\newblock In {\em The Twelfth International Conference on Learning Representations}, 2023.

\bibitem{katz1996synaptic}
Larry~C Katz and Carla~J Shatz.
\newblock Synaptic activity and the construction of cortical circuits.
\newblock {\em Science}, 274(5290):1133--1138, 1996.

\bibitem{kehayas2015dissonant}
Vassilis Kehayas and Anthony Holtmaat.
\newblock Dissonant synapses shall be punished.
\newblock {\em Neuron}, 87(2):245--247, 2015.

\bibitem{kembay2025quantitative}
Assel Kembay, Karina Aguilar, and Jason Eshraghian.
\newblock A quantitative analysis of catastrophic forgetting in quantized spiking neural networks.
\newblock In {\em 2025 IEEE International Symposium on Circuits and Systems (ISCAS)}, pages 1--5. IEEE, 2025.

\bibitem{kingma2014adam}
Diederik~P Kingma and Jimmy Ba.
\newblock Adam: A method for stochastic optimization.
\newblock {\em ArXiv preprint arXiv:1412.6980}, 2014.

\bibitem{krizhevsky2009learning}
Alex Krizhevsky, Geoffrey Hinton, et~al.
\newblock Learning multiple layers of features from tiny images.
\newblock 2009.

\bibitem{krizhevsky2012imagenet}
Alex Krizhevsky, Ilya Sutskever, and Geoffrey~E Hinton.
\newblock Imagenet classification with deep convolutional neural networks.
\newblock {\em Advances in Neural Information Processing Systems}, 25, 2012.

\bibitem{krueger1993neuronal}
JAMES~M KRUEGER and FERENC OB{\"A}L~JR.
\newblock A neuronal group theory of sleep function.
\newblock {\em Journal of Sleep Research}, 2(2):63--69, 1993.

\bibitem{lapicque2007quantitative}
Louis Lapicque.
\newblock Quantitative investigations of electrical nerve excitation treated as polarization. 1907.
\newblock {\em Biological Cybernetics}, 97(5-6):341--349, 2007.

\bibitem{lewicki2002efficient}
Michael~S Lewicki.
\newblock Efficient coding of natural sounds.
\newblock {\em Nature Neuroscience}, 5(4):356--363, 2002.

\bibitem{li2017cifar10}
Hongmin Li, Hanchao Liu, Xiangyang Ji, Guoqi Li, and Luping Shi.
\newblock Cifar10-dvs: an event-stream dataset for object classification.
\newblock {\em Frontiers in Neuroscience}, 11:309, 2017.

\bibitem{liu2021sstdp}
Fangxin Liu, Wenbo Zhao, Yongbiao Chen, Zongwu Wang, Tao Yang, and Li~Jiang.
\newblock Sstdp: Supervised spike timing dependent plasticity for efficient spiking neural network training.
\newblock {\em Frontiers in Neuroscience}, 15:756876, 2021.

\bibitem{loshchilov2016sgdr}
Ilya Loshchilov and Frank Hutter.
\newblock Sgdr: Stochastic gradient descent with warm restarts.
\newblock {\em ArXiv preprint arXiv:1608.03983}, 2016.

\bibitem{maass1997networks}
Wolfgang Maass.
\newblock Networks of spiking neurons: the third generation of neural network models.
\newblock {\em Neural Networks}, 10(9):1659--1671, 1997.

\bibitem{malyshev2013energy}
Aleksey Malyshev, Tatjana Tchumatchenko, Stanislav Volgushev, and Maxim Volgushev.
\newblock Energy-efficient encoding by shifting spikes in neocortical neurons.
\newblock {\em European Journal of Neuroscience}, 38(8):3181--3188, 2013.

\bibitem{markram1997regulation}
Henry Markram, Joachim L{\"u}bke, Michael Frotscher, and Bert Sakmann.
\newblock Regulation of synaptic efficacy by coincidence of postsynaptic {APs} and {EPSPs}.
\newblock {\em Science}, 275(5297):213--215, 1997.

\bibitem{florian2007reinforcement}
Naoki Masuda and Hiroshi Kori.
\newblock Formation of feedforward networks and frequency synchrony by spike-timing-dependent plasticity.
\newblock {\em Journal of Computational Neuroscience}, 22:327--345, 2007.

\bibitem{neftci2019surrogate}
Emre~O Neftci, Hesham Mostafa, and Friedemann Zenke.
\newblock Surrogate gradient learning in spiking neural networks: Bringing the power of gradient-based optimization to spiking neural networks.
\newblock {\em IEEE Signal Processing Magazine}, 36(6):51--63, 2019.

\bibitem{olshausen1996emergence}
Bruno~A Olshausen and David~J Field.
\newblock Emergence of simple-cell receptive field properties by learning a sparse code for natural images.
\newblock {\em Nature}, 381(6583):607--609, 1996.

\bibitem{paszke2019pytorch}
A~Paszke.
\newblock Pytorch: An imperative style, high-performance deep learning library.
\newblock {\em ArXiv preprint arXiv:1912.01703}, 2019.

\bibitem{patel2023local}
Adeetya Patel, Michael Eickenberg, and Eugene Belilovsky.
\newblock Local learning with neuron groups.
\newblock {\em arXiv preprint arXiv:2301.07635}, 2023.

\bibitem{table9}
Nicolas Perez-Nieves and Dan Goodman.
\newblock Sparse spiking gradient descent.
\newblock {\em Advances in Neural Information Processing Systems}, 34:11795--11808, 2021.

\bibitem{bib49}
Nicolas Perez-Nieves, Vincent~CH Leung, Pier~Luigi Dragotti, and Dan~FM Goodman.
\newblock Neural heterogeneity promotes robust learning.
\newblock {\em Nature Communications}, 12(1):5791, 2021.

\bibitem{pfeiffer2018deep}
Michael Pfeiffer and Thomas Pfeil.
\newblock Deep learning with spiking neurons: Opportunities and challenges.
\newblock {\em Frontiers in Neuroscience}, 12:409662, 2018.

\bibitem{pfister2006triplets}
Jean-Pascal Pfister and Wulfram Gerstner.
\newblock Triplets of spikes in a model of spike timing-dependent plasticity.
\newblock {\em Journal of Neuroscience}, 26(38):9673--9682, 2006.

\bibitem{popovych2013self}
Oleksandr~V Popovych, Serhiy Yanchuk, and Peter~A Tass.
\newblock Self-organized noise resistance of oscillatory neural networks with spike timing-dependent plasticity.
\newblock {\em Scientific Reports}, 3(1):2926, 2013.

\bibitem{quintana2022bio}
Fernando~M Quintana, Fernando Perez-Pena, and Pedro~L Galindo.
\newblock Bio-plausible digital implementation of a reward modulated {STDP} synapse.
\newblock {\em Neural Computing and Applications}, 34(18):15649--15660, 2022.

\bibitem{rathi2020enabling}
Nitin Rathi, Gopalakrishnan Srinivasan, Priyadarshini Panda, and Kaushik Roy.
\newblock Enabling deep spiking neural networks with hybrid conversion and spike timing dependent backpropagation.
\newblock {\em ArXiv preprint arXiv:2005.01807}, 2020.

\bibitem{bib50}
Julian Rossbroich, Julia Gygax, and Friedemann Zenke.
\newblock Fluctuation-driven initialization for spiking neural network training.
\newblock {\em Neuromorphic Computing and Engineering}, 2(4):044016, 2022.

\bibitem{saranirad2024cdna}
Vahid Saranirad, Shirin Dora, Thomas~Martin McGinnity, and Damien Coyle.
\newblock Cdna-snn: A new spiking neural network for pattern classification using neuronal assemblies.
\newblock {\em IEEE Transactions on Neural Networks and Learning Systems}, 2024.

\bibitem{sengupta2019going}
Abhronil Sengupta, Yuting Ye, Robert Wang, Chiao Liu, and Kaushik Roy.
\newblock Going deeper in spiking neural networks: {VGG} and residual architectures.
\newblock {\em Frontiers in Neuroscience}, 13:95, 2019.

\bibitem{sharifshazileh2021electronic}
Mohammadali Sharifshazileh, Karla Burelo, Johannes Sarnthein, and Giacomo Indiveri.
\newblock An electronic neuromorphic system for real-time detection of high frequency oscillations ({HFO}) in intracranial eeg.
\newblock {\em Nature Communications}, 12(1):3095, 2021.

\bibitem{shi2024spikingresformer}
Xinyu Shi, Zecheng Hao, and Zhaofei Yu.
\newblock Spikingresformer: Bridging {ResNet} and vision transformer in spiking neural networks.
\newblock In {\em Proceedings of the IEEE/CVF Conference on Computer Vision and Pattern Recognition}, pages 5610--5619, 2024.

\bibitem{singer1999neuronal}
Wolf Singer.
\newblock Neuronal synchrony: a versatile code for the definition of relations?
\newblock {\em Neuron}, 24(1):49--65, 1999.

\bibitem{song2000competitive}
Sen Song, Kenneth~D Miller, and Larry~F Abbott.
\newblock Competitive hebbian learning through spike-timing-dependent synaptic plasticity.
\newblock {\em Nature Neuroscience}, 3(9):919--926, 2000.

\bibitem{stuijt2021mubrain}
Jan Stuijt, Manolis Sifalakis, Amirreza Yousefzadeh, and Federico Corradi.
\newblock $\mu$brain: An event-driven and fully synthesizable architecture for spiking neural networks.
\newblock {\em Frontiers in Neuroscience}, 15:664208, 2021.

\bibitem{tavanaei2019deep}
Amirhossein Tavanaei, Masoud Ghodrati, Saeed~Reza Kheradpisheh, Timoth{\'e}e Masquelier, and Anthony Maida.
\newblock Deep learning in spiking neural networks.
\newblock {\em Neural Networks}, 111:47--63, 2019.

\bibitem{wu2018spatio}
Yujie Wu, Lei Deng, Guoqi Li, Jun Zhu, and Luping Shi.
\newblock Spatio-temporal backpropagation for training high-performance spiking neural networks.
\newblock {\em Frontiers in Neuroscience}, 12:331, 2018.

\bibitem{xiao2017fashion}
Han Xiao, Kashif Rasul, and Roland Vollgraf.
\newblock Fashion-mnist: a novel image dataset for benchmarking machine learning algorithms.
\newblock {\em ArXiv preprint arXiv:1708.07747}, 2017.

\bibitem{yao2024spike}
Man Yao, Jiakui Hu, Zhaokun Zhou, Li~Yuan, Yonghong Tian, Bo~Xu, and Guoqi Li.
\newblock Spike-driven transformer.
\newblock {\em Advances in Neural Information Processing Systems}, 36, 2024.

\bibitem{bib48}
Bojian Yin, Federico Corradi, and Sander~M Boht{\'e}.
\newblock Accurate and efficient time-domain classification with adaptive spiking recurrent neural networks.
\newblock {\em Nature Machine Intelligence}, 3(10):905--913, 2021.

\bibitem{zenke2018superspike}
Friedemann Zenke and Surya Ganguli.
\newblock Superspike: Supervised learning in multilayer spiking neural networks.
\newblock {\em Neural Computation}, 30(6):1514--1541, 2018.

\bibitem{zenke2021remarkable}
Friedemann Zenke and Tim~P Vogels.
\newblock The remarkable robustness of surrogate gradient learning for instilling complex function in spiking neural networks.
\newblock {\em Neural Computation}, 33(4):899--925, 2021.

\bibitem{table5}
Dongcheng Zhao, Yi~Zeng, Tielin Zhang, Mengting Shi, and Feifei Zhao.
\newblock {GLSNN: A} multi-layer spiking neural network based on global feedback alignment and local {STDP} plasticity.
\newblock {\em Frontiers in Computational Neuroscience}, 14:576841, 2020.

\bibitem{exp1}
Hanle Zheng, Zhong Zheng, Rui Hu, Bo~Xiao, Yujie Wu, Fangwen Yu, Xue Liu, Guoqi Li, and Lei Deng.
\newblock Temporal dendritic heterogeneity incorporated with spiking neural networks for learning multi-timescale dynamics.
\newblock {\em Nature Communications}, 15(1):277, 2024.

\bibitem{zhou2023spikingformer}
Chenlin Zhou, Liutao Yu, Zhaokun Zhou, Zhengyu Ma, Han Zhang, Huihui Zhou, and Yonghong Tian.
\newblock Spikingformer: Spike-driven residual learning for transformer-based spiking neural network.
\newblock {\em ArXiv preprint arXiv:2304.11954}, 2023.

\bibitem{zhou2022spikformer}
Zhaokun Zhou, Yuesheng Zhu, Chao He, Yaowei Wang, Shuicheng Yan, Yonghong Tian, and Li~Yuan.
\newblock Spikformer: When spiking neural network meets transformer.
\newblock {\em ArXiv preprint arXiv:2209.15425}, 2022.

\end{thebibliography}










\clearpage
\appendix
\label{sup}

\section{Supplementary Note 1: Experimental Setup and Hyperparameters}
\label{sup1}
\textbf{Hardware Configuration}
All experiments were conducted using NVIDIA RTX 4090 Laptop and Tesla V100 GPUs.

\textbf{Single Hidden Layer SNN Models}
For all single hidden layer SNN models evaluated on FMNIST, N-MNIST, and CIFAR-10, we adopted the same architecture and hyperparameters. Models were trained for 100 epochs using the Adam optimizer~\cite{kingma2014adam}, with cosine annealing~\cite{loshchilov2016sgdr} applied to both the learning rate and all SSDP parameters. No data augmentation was employed during training. The exponential decay variant of SSDP was used throughout, with parameters set to $A_{+} = 0.02$, $A_{-} = 0.02$, $\tau_{+} = 20.0$, and $\tau_{-} = 20.0$.

\textbf{DHSNN Models on SHD and SSC}
For the SHD dataset, we used a single-layer DHSNN with SSDP implemented via the Gaussian variant. For SSC, we adopted a two-layer DHSNN configuration. Both models were trained using Adam optimizer and cosine-annealed learning rate schedules for 100 (SHD) and 150 (SSC) epochs, respectively, starting SSDP updates from 10 epochs. The learning rate was set to $5 \times 10^{-4}$. The SSDP parameters for both tasks were $A_{+} = 1.5 \times 10^{-4}$, $A_{-} = 5 \times 10^{-5}$, $A_{\text{-}} = 5 \times 10^{-5}$, and $\sigma = 1.0$.

\textbf{Spiking-ResNet18 on CIFAR-10 and CIFAR-100}
For experiments using Spiking-ResNet18 architectures on CIFAR-10 and CIFAR-100, we used identical model configurations across both datasets. SSDP was configured with the exponential variant and parameters set to $A_{+} = 5 \times 10^{-4}$, $A_{-} = 5 \times 10^{-4}$, $\tau_{+} = 20.0$, and $\tau_{-} = 20.0$. Training was performed using the SGD optimizer~\cite{bottou2010large} with a learning rate schedule defined as MultiStepLR (optimizer, milestones=[45, 90, 120], gamma=0.2).

\textbf{Spiking-transformer with SSDP on CIFAR-10, CIFAR-100, and ImageNet}
For SSD-ViT models evaluated on CIFAR-10, CIFAR-100, and ImageNet, we trained for 600 epochs using the AdamW optimizer with a base learning rate of $5 \times 10^{-4}$. Cosine annealing schedules were applied to both the learning rate and SSDP parameters, with SSDP training initiated after 10 epochs. On the CIFAR10-DVS dataset, we used the same configuration but trained for 100 epochs. Specific SSDP hyperparameter settings for these experiments are provided in the main text.

\section{Supplementary Note 2: Neuron Model and Exponential SSDP Variant (Methods Detail)}
\label{sup2}

\subsection{SNNs neuron dynamics}

In our single hidden layer SNNs model, the temporal behavior of the spiking neurons is captured using a recurrent architecture that emulates the leaky integration \cite{lapicque2007quantitative} observed in biological systems. At each discrete time step, the membrane potential \( m(t) \) is updated according to an exponential decay mechanism defined by
\begin{equation}
\alpha = \exp\left(-\frac{\Delta t}{\tau_m}\right),
\end{equation}
where \( \tau_m \) denotes the membrane time constant and \( \Delta t \) is the time interval between updates. The previous membrane potential \( m(t-1) \) is attenuated by \( \alpha \), while the current synaptic input, comprising both external stimuli and recurrent feedback, is integrated into the updated state.

To further approximate biological processes, the model incorporates a dendritic integration mechanism \cite{exp1}. The dendritic input \( d_{\text{input}}(t) \) is computed as:
\begin{equation}
d_{\text{input}}(t) = \sigma(\tau_n) \cdot d_{\text{input}}(t-1) + \left[1 - \sigma(\tau_n)\right] \cdot I(t)
\end{equation}
where \( \tau_n \) is a learnable dendritic time constant, \( \sigma(\cdot) \) is the sigmoid function, and \( I(t) \) represents the combined synaptic current. The overall update of the membrane potential is then given by
\begin{equation}
m(t) = \alpha \cdot m(t-1) + (1 - \alpha) \cdot d_{\text{input}}(t)
\end{equation}
spike generation is triggered when the membrane potential exceeds a fixed threshold \( v_{\text{th}} \). Spiking neurons are non-differentiable due to their binary firing behavior, so we adopt a custom activation function: it behaves as a hard threshold in the forward pass and uses a multi-Gaussian surrogate gradient for backpropagation, enabling gradient-based optimization.

We implement the neuron model in PyTorch \cite{paszke2019pytorch}, combining leaky membrane dynamics, dendritic integration, and surrogate gradients. This design enables the simulation of temporally precise spiking behavior and supports the study of plasticity in neuromorphic systems.

\subsection{Exponential SSDP}

\textbf{Learnable scalars.}
\(A_{+}\) (potentiation scale), \(A_{-}\) (depression scale),
\(\tau_{+}\) (causal time constant), \(\tau_{-}\) (anti-causal time constant).

\noindent \textbf{Inputs.}
The module receives per-sample spike–presence indicators
\[
Q_{b,i}\in\{0,1\}\quad(\text{pre/input}),\qquad
P_{b,j}\in\{0,1\}\quad(\text{post/output}),
\]
arranged as tensors \(\texttt{pre\_spike}\in\mathbb{R}^{B\times N_{\mathrm{in}}}\),
\(\texttt{post\_spike}\in\mathbb{R}^{B\times N_{\mathrm{out}}}\).
From these we form a \emph{synchrony gate} by a broadcasted outer product
\[
\lambda_{b,ji}\;=\;P_{b,j}\,Q_{b,i}\in\{0,1\},
\]
which equals \(1\) iff both neurons spike at least once within sample \(b\).
The module also receives a nonnegative per-sample lag \(\Delta t_b\) as
\(\texttt{delta\_t}\in\mathbb{R}^{B\times 1}\) (shared across all pairs for that sample).

\noindent \textbf{Update rule.}
With the above gate and a decaying exponential in the lag, the batch update implemented by the module is
\begin{align}
\Delta W_{\text{pot}}
&= \frac{A_{+}}{B}\sum_{b=1}^{B}\; e^{-\Delta t_b/\tau_{+}}\;\lambda_{b}
\;\;\in\mathbb{R}^{N_{\mathrm{out}}\times N_{\mathrm{in}}}\\
\Delta W_{\text{dep}}
&= \frac{A_{-}}{B}\sum_{b=1}^{B}\; e^{-\Delta t_b/(\tau_{+}\tau_{-})}\;\lambda_{b}\\
\Delta W &= \Delta W_{\text{pot}} - \Delta W_{\text{dep}},\qquad
W \leftarrow \mathrm{clip}\!\left(W+\Delta W,\,-1,\,1\right)
\end{align}
Here \(\lambda_b = P_b\,Q_b^{\top}\) is the \((N_{\mathrm{out}}\!\times N_{\mathrm{in}})\) synchrony mask for sample \(b\), and the exponential factors are scalars broadcast to that matrix.
\emph{As implemented in code}, the depression kernel uses \(e^{-\Delta t_b/(\tau_{+}\tau_{-})}\) (because \(\Delta t\) is first normalised by \(\tau_{+}\) and then divided by \(\tau_{-}\) before exponentiation); thus both potentiation and depression are gated by the same \(\lambda\) but decay with different effective time scales.

\noindent \textbf{Scope.}
This exponential variant was only used in tiny sanity tests; all main-text results adopt the symmetric Gaussian SSDP described in the Methods.


\subsection{Ablation Studies}

\subsubsection{Different SSDP Variants in
SNN-Transformer}

We ablate the synchrony update to test whether SSDP is functionally necessary in Table~\ref{tab:ssdp-2}. On CIFAR-100 with the SNN-Transformer, removing SSDP reduces top-1 accuracy relative to the same model trained with SSDP. And change Gaussian kernel to Exponential kernel will also reduce the gain.
With the symmetric Gaussian window we adopt, accuracy improves from $78.73\%$ to $79.48\%$ under a matched schedule.

\paragraph{Why Gaussian.}
SSDP is intended to reward \emph{near-coincident} group activity rather than precise spike ordering.
A symmetric Gaussian kernel implements this directly: (i) it maximises the update at zero lag and decays smoothly on both sides, tolerating small timing jitter; (ii) it avoids introducing an arbitrary lead/lag preference, aligning with the rate-controlled evidence of a \emph{symmetric} zero-lag peak Fig.~\ref{fig:ssdp_synchrony}(a–b); and (iii) it yields bounded, smooth contributions that stabilise late training when combined with clipping.
These properties agree with our empirical findings: Fig.~\ref{fig:combined} show that with SSDP the representation becomes more organised along a few shared directions while spike timing becomes more reliable (tails shrink without forcing rigidity).
In contrast, an asymmetric, order-sensitive decay can overemphasise fine timing differences and is less robust under event noise or frame-to-event conversion artifacts.

\begin{table*}[htbp]
\centering
\begin{minipage}{0.48\linewidth}
\centering
\caption{Top-1 classification accuracy (\%) on CIFAR-100 with different SSDP variants in SNN-Transformer.}
\label{tab:ssdp-2}
\rowcolors{2}{white}{gray!10}
\begin{adjustbox}{max width=\linewidth}
{\footnotesize
\begin{tabular}{lcc}
\hline
\textbf{Model Variant} & \textbf{SSDP Type} & \textbf{Accuracy (\%)} \\
\hline
Baseline (No SSDP) & --- & 78.73 \\
+ SSDP & Exponential decay & 79.12 (+0.39) \\
+ SSDP & Gaussian decay & \textbf{79.48 (+0.75)} \\
\hline
\end{tabular}
}
\end{adjustbox}
\end{minipage}
\hfill
\begin{minipage}{0.48\linewidth}
\centering
\caption{Performance under different $A_{\text{+}}$ and $A_{\text{-}}$ settings in the SSDP rule.}
\label{tab:ssdp-params}
\rowcolors{2}{white}{gray!10}
\begin{adjustbox}{max width=\linewidth}
{\footnotesize
\begin{tabular}{ccc}
\hline
$\boldsymbol{A_{\text{+}}}$ & $\boldsymbol{A_{\text{-}}}$ & \textbf{Accuracy (\%)} \\
\hline
$1.5 \times 10^{-4}$ & $5.0 \times 10^{-5}$ & 79.48 \\
$1.0 \times 10^{-4}$ & $1.0 \times 10^{-4}$ & 79.03 \\
$5.0 \times 10^{-5}$ & $1.5 \times 10^{-4}$ & 79.07 \\
$1.0 \times 10^{-2}$ & $1.0 \times 10^{-2}$ & Divergent \\
\hline
\end{tabular}
}
\end{adjustbox}
\end{minipage}
\end{table*}

\subsubsection{Optimal hyperparameter tuning balances synchrony-driven plasticity}

We sweep the SSDP strength on the SNN–Transformer (CIFAR-100) over a log-spaced grid of the potentiation scale \(A_{+}\) and the depression \(A_{-}\) under a matched schedule and fixed seeds; all other hyperparameters are held constant and each run uses the same epoch budget. Validation top-1 is summarised in Table~\ref{tab:ssdp-params}. The accuracy landscape forms a ridge rather than a single sharp optimum. When \(A_{+}\) is too small, the SSDP contribution is negligible and accuracy regresses toward the no-SSDP baseline. Performance is maximised around \((A_{+},A_{-})=(1.5{\times}10^{-4},\,5{\times}10^{-5})\) used in our main models; pushing either parameter much higher quickly destabilises training -- once \(A_{+}\) approaches the \(10^{-3}\) scale together with proportionally large \(A_{-}\), runs fail to converge (divergent loss and accuracy collapse). Empirically, settings with \(A_{+}>A_{-}\) work best: near-synchronous pairs receive net potentiation while asynchronous pairs remain weakly depressed. If \(A_{-}\) approaches \(A_{+}\), the net effect becomes overly suppressive and learning stalls; if \(A_{+}\) is made far larger, early or random coincidences can trigger runaway growth. This ridge location -- \(A_{+}\) stronger than \(A_{-}\) -- is consistent with our rate-controlled finding that same-time co-activation is positively enriched (Fig.~\ref{fig:ssdp_synchrony}(a–b). At settings on the ridge, we also observe the representation to spread mainly along a few shared axes (PCA) and spike-time jitter to tighten without forcing rigidity (Fig.~\ref{fig:combined}); off-ridge, these effects weaken or training becomes unstable. Practically, we recommend starting from \((A_{+},A_{-})=(1.5{\times}10^{-4},\,5{\times}10^{-5})\), tuning on a log scale while keeping \(0<A_{-}<A_{+}\). If gains vanish, increase \(A_{+}\) one notch at a similar ratio; if training oscillates or diverges, first reduce \(A_{+}\) and then \(A_{-}\). 

A related biological observation is that correlated inputs tend to be stabilised while uncorrelated inputs are weakened \cite{song2000competitive,katz1996synaptic}. 
For example, temporally coordinated synaptic activity can be maintained, whereas asynchronous inputs may undergo activity-dependent weakening in specific preparations \cite{kehayas2015dissonant}. This picture is consistent with our strength calibration: setting $A_{+}>A_{-}$ lets near-synchronous events produce net potentiation while asynchronous pairs remain weakly depressed.

\section{Supplementary Note 3: Integration into Single Hidden Layer SNNs}
\label{sup3}
\begin{table}[htbp]
\centering
\caption{ Classification accuracy (Top-1) based on SNNs. The proposed model consists of a single hidden layer and a readout classifier, without any convolutional components.}
\label{tab:mnist-results}

\rowcolors{2}{white}{gray!10}
\begin{adjustbox}{max width=\linewidth}
\resizebox{0.67\linewidth}{!}{
\begin{tabular}{llcc}
\hline
\textbf{Dataset} & \textbf{Learning rule} & \textbf{Conv. structure} & \textbf{Accuracy (\%)} \\
\hline
\textbf{Fashion-MNIST} & Sym-STDP \cite{hao2020biologically} & \textcolor{red}{\ding{55}} & 85.3 \\
 & GLSNN \cite{table5} & \textcolor{red}{\ding{55}} & 89.1 \\
 & STB-STDP \cite{table1} & \textcolor{green}{\ding{51}} & 87.0 \\
 & R-STDP \cite{quintana2022bio} & \textcolor{green}{\ding{51}} & 83.26 \\
 & SSTDP \cite{liu2021sstdp} & \textcolor{green}{\ding{51}} & 85.16 \\
 & S2-STDP \cite{goupy2024neuronal} & \textcolor{green}{\ding{51}} & 85.88 \\
 & \textbf{Proposed (1 hidden layer)} & \textcolor{red}{\ding{55}} & \textbf{89.37$_{\pm 0.14}$} \\
 \hline
\textbf{CIFAR-10} & STB-STDP & \textcolor{green}{\ding{51}} & 32.95 \\
 & R-STDP & \textcolor{green}{\ding{51}} & 51.74 \\
 & S2-STDP & \textcolor{green}{\ding{51}} & 61.80 \\
 & SSTDP & \textcolor{green}{\ding{51}} & 60.80 \\
 & NCG-S2-STDP \cite{goupy2024neuronal} & \textcolor{green}{\ding{51}} & 66.41 \\
 & \textbf{Proposed (1 hidden layer)} & \textcolor{red}{\ding{55}} & \textbf{52.15$_{\pm0.23}$} \\
\hline
\textbf{N-MNIST} & STDBP & \textcolor{red}{\ding{55}} & 98.74 \\
 & Synaptic Kernel Inverse \cite{cohen2016skimming} & \textcolor{red}{\ding{55}} & 92.87 \\
 & BP-SNN \cite{table9}& \textcolor{red}{\ding{55}} & 92.7 \\
 & STDP+CDNA-SNN \cite{saranirad2024cdna} & \textcolor{red}{\ding{55}} & 98.43 \\
 & \textbf{Proposed (1 hidden layer)} & \textcolor{red}{\ding{55}} & \textbf{96.54$_{\pm 0.11}$} \\

\hline
\end{tabular}
}
\end{adjustbox}
\end{table}
\begin{table}[htbp]
    \centering
    \begin{minipage}[t]{0.48\linewidth}
        \centering
        \caption{Classification accuracy (Top-1) based on Spiking-ResNet18 \cite{hu2021spiking} compared with ANN baselines.}
        \label{tab:spiking-resnet18}
        \rowcolors{2}{white}{gray!10}
        \resizebox{\linewidth}{!}{
        \begin{tabular}{lcc}
        \hline
        \textbf{Dataset} & \textbf{Model}  & \textbf{Accuracy (\%)} \\
        \hline
        \textbf{CIFAR-10} 
         & SpikeResNet18-STDP  & 76.71 \\
         & ResNet20 (ANN) \cite{sengupta2019going}  & 89.1 \\
         & STBP \cite{wu2018spatio}   & 89.53 \\
         & EIHL \cite{jiang2023adaptive}  & 90.25 \\
         & ResNet20 (ANN) \cite{han2020rmp}  & 91.47 \\ 
         & \textbf{Proposed}  & \textbf{91.37$_{\pm 0.37}$} \\
        \hline
        \textbf{CIFAR-100}  
         & SpikeResNet18-STDP & 32.66 \\
         & STBP & 58.48 \\
         & EIHL  & 58.63 \\
         & ResNet20 (ANN) \cite{han2020rmp}  & 68.72 \\ 
         & \textbf{Proposed}  & \textbf{60.97$_{\pm 0.42}$} \\
        \hline
        \end{tabular}
        }
    \end{minipage}%
    \hfill
    \begin{minipage}[t]{0.48\linewidth}
        \centering
        \caption{Extended evaluation metrics on CIFAR-100 using SNN-Transformer, with and without SSDP.}
        \label{tab:ssdp-cifar100-extended}
        \rowcolors{2}{white}{gray!10}
        \begin{adjustbox}{max width=\linewidth}
        {\footnotesize
        \begin{tabular}{lcc}
        \hline
        \textbf{Metric} & \textbf{Without SSDP} & \textbf{With SSDP} \\
        \hline
        Top-1 Accuracy        & 78.73\%        & \textbf{79.48\%} \\
        Top-5 Accuracy        & 93.64\%        & 93.58\% \\
        Throughput (it/s)     & 352.75         & 352.31 \\
        MACs                  & 1.226 G        & 1.226 G \\
        Parameters            & 10.83 M        & 10.83 M \\
        Avg SOPs              & 0.548 G        & 0.560 G \\
        Power (mJ/sample)     & 0.4936         & 0.5043 \\
        A/S Power Ratio       & 11.43          & 11.19 \\
        \hline
        \end{tabular}
        }
        \end{adjustbox}
    \end{minipage}
\end{table}
We first conducted preliminary evaluations of SSDP on a simple single-hidden-layer SNN without convolutional structures. Experimental results in Table~\ref{tab:mnist-results} showed that our model achieved an accuracy of 89.37\% on the static Fashion-MNIST dataset, comparable to certain deep SNN architectures and convolutional STDP-based models. On the more challenging CIFAR-10 dataset, our single-layer SSDP model reached an accuracy of 52.15\%, slightly outperforming the more complex convolutional R-STDP model but still behind other advanced STDP-based models. Subsequently, we evaluated our model on the event-based N-MNIST dataset; however, the performance did not surpass existing STDP-based methods. We provide an in-depth analysis of SSDP's performance on DVS datasets in the following subsection.


\section{Supplementary Note 4: Cross-architecture overview and design rationale}
\label{sup4}

\subsection{Cross-architecture overview \& design rationale}

Early SNNs were shallow and difficult to optimise due to binary activations and non-differentiable thresholds. Residual designs (e.g., spiking ResNet-style networks) alleviated vanishing gradients, yet training stability and representation quality remained sensitive to timing noise and depth. More recently, SNN-Transformers reintroduced global dependency modelling via spike-compatible self-attention. This evolution motivates a plasticity rule that (i) exploits short-window population synchrony, (ii) adds no inference-time cost, and (iii) scales to both convolutional blocks and attention stacks.

We integrate SSDP in a purely training-time manner, leaving the forward graph unchanged. In convolutional models (Spiking ResNet18), SSDP is attached after stage-end projection layers and/or at the classifier, where activity is sparser and co-activation carries task signal. In SNN-Transformers (e.g., Spikformer/SpikingResformer), SSDP is attached at the $1{\times}1$ projection after the last DSSA block in the final stage and at the linear classifier (see Fig.~\ref{fig:SSD-ViT}). Each hook records binary spike indicators and computes a batchwise outer-product update gated by synchrony; cost is $\mathcal{O}(C_{\text{out}}C_{\text{in}})$ per batch with clipping, and no state is kept at inference.

Across both families, the same implementation yields consistent accuracy gains under matched budgets (Main Results tables) and smoother optimisation (Fig.~\ref{fig:spike-train3}). Gains are larger at mid/late stages where synchrony is more task-aligned, which is consistent with the rate-controlled analyses (Fig.~\ref{fig:ssdp_synchrony}). Full per-layer placement ablations and complexity measurements are provided in Supplementary (\ref{sup6}).

\subsection{Architecture Details of SSDP-based SNN-Transformer}

This section presents an overview of the proposed \emph{SSD-ViT}. The architecture leverages time-step-based processing of spiking inputs alongside multi-head self-attention in the spiking domain to capture spatial and temporal dependencies.
\begin{figure}[!t]
    \centering
    \includegraphics[width=1.0\linewidth]{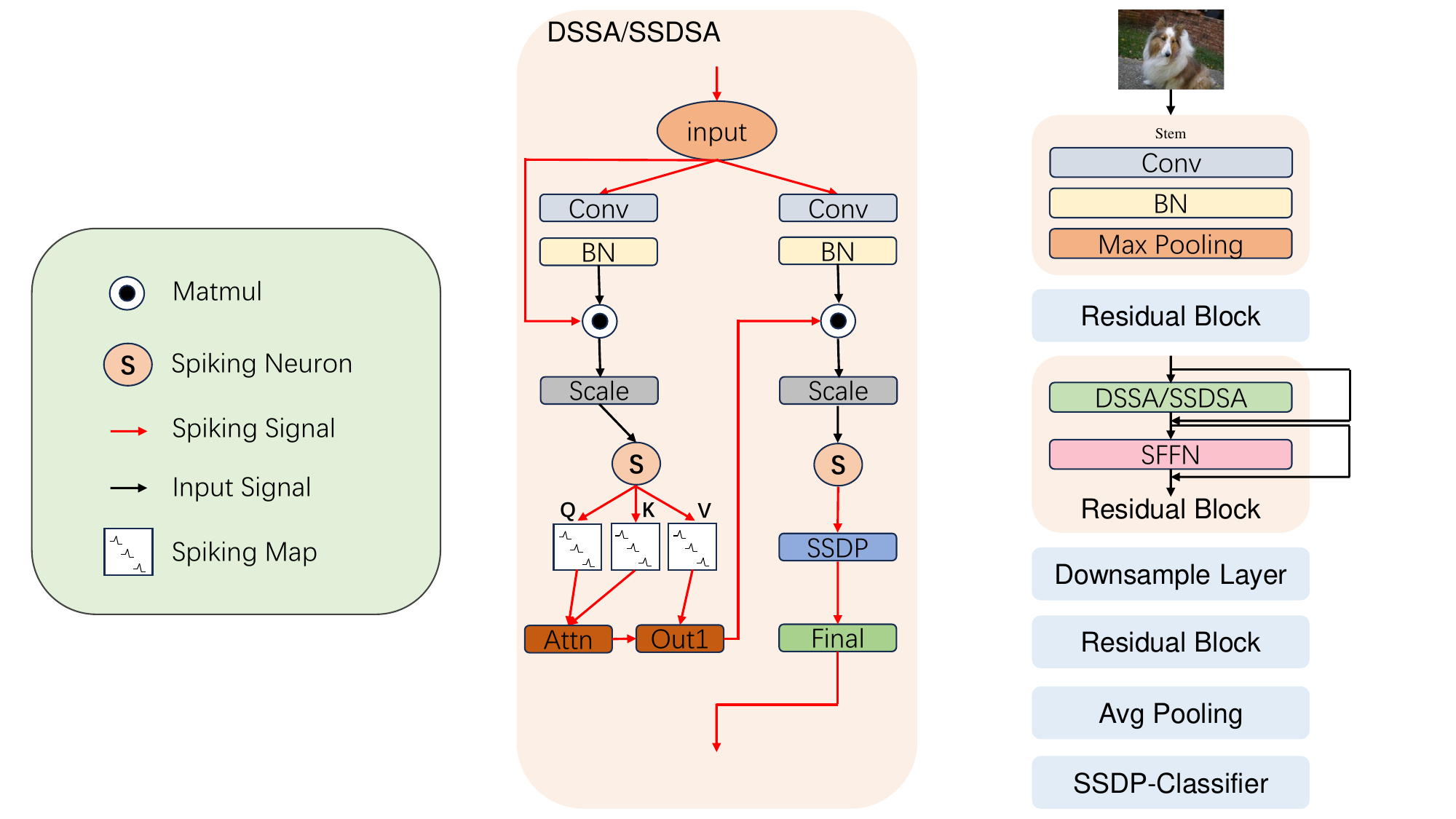}
    \caption{Overview of the Spiking Synchrony Dependent Self-Attention (SSDSA) and SSD-ViT architecture.}
    \label{fig:SSD-ViT}
\end{figure}
As illustrated in Fig.~\ref{fig:SSD-ViT}, the SSD-ViT processes an input spike tensor.
\(
\mathbf{X} \in \mathbb{R}^{T \times B \times C \times H \times W}
\)
through a series of components: 
\\
\textbf{Pulsed Input and Prologue.} The raw input spikes first undergo initial convolution and pooling, reducing spatial resolution and adjusting the channel dimension to produce an intermediate spiking feature map 
    \(\mathbf{X}\).
\\
\textbf{Spiking Synchrony Dependent Self-Attention (SSDSA).} 
    We propose SSDSA as an extension of Dual Spike Self-Attention (DSSA), augmented with SSDP mechanism. We make the input feature map be $\mathbf{X}\in \mathbb{R}^{T\times B\times C\times H\times W}$, 
    where $T$ denotes the number of discrete time steps, 
    $B$ is the batch size, 
    $C$ is the number of channels, 
    and $(H,W)$ indicates the spatial resolution. 
    We assume the network splits the channels into $n_{H}$ attention heads (i.e., $C$ is divisible by $n_{H}$). 
    Formally, SSDSA proceeds in two main steps: a \emph{spiking attention transformer} and a \emph{projection layer with spike-driven plasticity}.
    
    \paragraph{(a) Spiking Attention Transformer.} 
    First, we apply a spiking activation function (e.g., LIF) to $\mathbf{X}$:
    \begin{equation}
    X_{in} \;=\; \sigma_{in}\!\bigl(X\bigr)
    \end{equation}
    $\mathbf{Y} = \mathrm{BN}\bigl(\mathbf{W}(\mathbf{X}_{\mathrm{in}})\bigr)$, 
    splitting $\mathbf{Y}$ into two parts along the channel dimension,
    \(
    (\mathbf{Y}_1,\;\mathbf{Y}_2) = \mathrm{split}(\mathbf{Y}),
    \)
    so that each sub-tensor aligns with multi-head attention reshaping. 
    To obtain the spiking attention map, we compute
    \begin{equation}
    A \;=\; \sigma_{attn}\!\Bigl(
       Matmul\bigl(Y_1^T,\,X_{in}\bigr)\times \alpha_{1}
    \Bigr)
    \end{equation}
    \begin{equation}
    Z \;=\; \sigma_{out}\!\Bigl(
       Matmul\bigl(Y_2,\,A\bigr)\times \alpha_{2}
    \Bigr)
    \end{equation}
    where $\sigma_{\mathrm{attn}}(\cdot)$ and $\sigma_{\mathrm{out}}(\cdot)$ are spiking or nonlinear activations, 
    while $\mathrm{Matmul}(\cdot,\cdot)$ denotes a spiking matrix multiplication that respects the multi-head shape. 
    The terms $\alpha_{1},\alpha_{2}$ are scaling factors (often derived from neuron firing-rate statistics) to stabilize spiking operations.
    
    \paragraph{(b) Projection \& Residual Connection.}
    We then project $\mathbf{Z}$ back to the original channel dimension $C$ via 
    \begin{equation}
    \mathbf{O} \;=\; \mathrm{BN}\!\bigl(\mathbf{W}_{\mathrm{proj}}(\mathbf{Z})\bigr)\;+\;\mathbf{X}
    \end{equation}
    where $\mathbf{W}_{\mathrm{proj}}\in\mathbb{R}^{C\times C\times 1\times 1}$ is a $1\times1$ convolution, followed by a residual skip connection to the input $\mathbf{X}$. This yields the SSDSA output $\mathbf{O}\in \mathbb{R}^{T\times B\times C\times H\times W}$.
    
    \paragraph{(c) Synaptic Plasticity for the Projection Layer.}
    To further exploit spike timing correlations, SSDSA integrates SSDP on $\mathbf{W}_{\mathrm{proj}}$. 
    Concretely, let $\mathbf{S}_{\mathrm{pre}}\in\{0,1\}$ and $\mathbf{S}_{\mathrm{post}}\in\{0,1\}$ be binary indicators capturing whether each input or output neuron emits a spike, 
    and $t_{\mathrm{pre},b,i}$, $t_{\mathrm{post},b,j}$ be their respective firing times (e.g., the first time a spike occurs). 
    We define:

    \begin{equation}
    \Delta t_{b,j,i} \;=\;
    \bigl|\,
     t_{\mathrm{post},b,j}\;-\;t_{\mathrm{pre},b,i}
    \bigr|
    \end{equation}
    \begin{equation}
    synchronized_{b,j,i}
    \;=\;
    S_{pre,b,i}\,\times\,S_{post,b,j}
    \end{equation}
    furthermore, let
    \begin{equation}
    g
    \;=\;
    \exp\!\Bigl(
       -\,\tfrac{\Delta t_{b,j,i}^{2}}{2\,\sigma^2}
    \Bigr)
    \end{equation}

    we denote by \( t_{\mathrm{pre},b,i} \) the first-spike time for the \( i \)-th pre-synaptic neuron within the \( b \)-th sample (or mini-batch element). In other words, for each data sample \( b \) and each input neuron \( i \), \( t_{\mathrm{pre},b,i} \) is the time step at which that neuron emits its first spike. Similarly, \( t_{\mathrm{post},b,j} \) refers to the first-spike time for the \( j \)-th post-synaptic neuron in the \( b \)-th sample. \( \text{pre\_spike}_{b,i}(t) \) and \( \text{post\_spike}_{b,j}(t) \) indicate whether the \( i \)-th (or \( j \)-th) neuron fires a spike at time \( t \) in the \( b \)-th sample. 
    Then, for each batch $b$, the update to $\mathbf{W}_{\mathrm{proj}}$ follows:

    \begin{align}
    \Delta \mathbf{W}_{j,i}
    &=
    mean_{b}\Bigl[
       A_{+}\,synchronized_{b,j,i}\;\gamma
       \nonumber \\[3pt]
    &\quad
       -\,A_{-}\,
       \bigl(1-synchronized_{b,j,i}\bigr)\,\gamma
    \Bigr]
    \\[4pt]
    W_{proj}
    &\;\leftarrow\;W_{proj} + \Delta W
    \end{align}
    where $\Delta W \in R^{C\times C}$ matches the projection-layer dimensionality.
    The hyperparameters, \( A_{+} \) and \( A_{\text{-}} \), which respectively govern the magnitudes of synaptic potentiation and synaptic depression. These parameters are empirically determined through experimental tuning to optimize the model's performance, and $\sigma$ controls the Gaussian sensitivity window over spike-time differences. 
    This spike-driven update effectively merges conventional gradient-based learning with temporal self-organization, 
    encouraging channels that fire synchronously (within short time lags) to strengthen their synaptic connections.

Thus, SSDSA incorporates multi-head spiking self-attention to model cross-channel dependencies, 
and refines its final projection layer via spike-based synaptic plasticity. 
Then, the network gains an additional mechanism for harnessing temporal correlations in neural activations, which improves spatiotemporal feature representations compared to purely gradient-based alternatives.
\\
 \textbf{Spiking Feed-Forward Network (SFFN).} Following the attention block, the network applies a feed-forward sub-module with grouped convolutions and spiking activations:
    \begin{equation}
    Y \;=\;
    \text{Down}\!\Bigl(\,\text{Up}(X) \;+\; 
    \sum_{i=1}^{num\_conv} \text{Conv3x3}\bigl(X\bigr)\Bigr) 
    \;+\; X
    \end{equation}
    here, \(\text{Up}\) and \(\text{Down}\) represent \(1\times 1\) convolutions interleaved with LIF and BN operations, while the interior \(\text{Conv3x3}\) blocks enhance local feature extraction.
\\
 \textbf{DownsampleLayer.} If necessary, the spatial resolution is further reduced via down-sampling (stride-2 convolutions) to balance feature richness and computational efficiency in deeper layers.
\\
 \textbf{Global Pooling and Classification.} Ultimately, an adaptive average pooling condenses the spiking feature maps across spatial dimensions. Formally, we extract:
    \begin{equation}
    h^{(t)} \;=\; pool\bigl(x^{(t)}\bigr)
    \quad \forall\, t=1,\dots,T
    \end{equation}
    and aggregate them over time:
    \begin{equation}
    \hat{y} 
    \;=\; 
    \tfrac{1}{T} \sum_{t=1}^{T} 
    W_{\text{cls}}\,h^{(t)}
    \end{equation}
    where \(\mathbf{W}_{\text{cls}}\) is the learnable weight of the classifier. The final prediction \(\hat{\mathbf{y}}\) may be interpreted as the mean spiking response across all time steps.

\noindent According to the detailed architecture in Table~\ref{tab:arch}, by operating in the spiking domain at each stage, our SSD-ViT is designed to exploit temporal spike patterns more effectively while retaining the powerful representational benefits of multi-head attention. Subsequently, we couple the primary supervised training objective with an additional unsupervised plasticity rule (SSDP), which refines the model weights based on spike-time synchronization, further improving the temporal sensitivity and representation robustness.

\begin{table}[!t]
\caption{Overview of the \textit{SSD-ViT} architecture. Each DSSA (or SSDSA) includes spiking attention (matmul1, matmul2) and LIF activations; each SFFN block comprises up/down 1$\times$1 conv and spiking Conv3$\times$3 with BN.}
\label{tab:arch}

\centering
\begin{adjustbox}{max width=\linewidth}
\begin{tabular}{|l|l|p{0.62\linewidth}|}
\hline
\textbf{Module} & \textbf{Sub-layer} & \textbf{Details} \\
\hline

\textbf{Prologue} 
  & Conv2d(3$\to$64, 3$\times$3), BN
  & Initial conv + BN, output shape: [B, 64, H, W]. \\
\hline

\textbf{Stage (0)} 
  & \textit{DSSA} 
  & LIF($\tau=2.0$); Conv2d(64$\to$128, 4$\times$4, stride=4), BN; SpikingMatmul $\times$2; Wproj: Conv1$\times$1(64$\to$64), BN. \\
  & \textit{SFFN} 
  & Up: LIF + Conv1$\times$1(64$\to$256), BN; Conv (LIF + Conv3$\times$3(256$\to$256, groups=4) + BN); Down: LIF + Conv1$\times$1(256$\to$64), BN. \\
\hline

\textbf{Stage (1)} 
  & \textit{Downsample} 
  & Conv3$\times$3(64$\to$192, stride=2), BN, LIF \\
  & \textit{DSSA} 
  & LIF; Conv2d(192$\to$384, 2$\times$2, stride=2), BN; matmul1, matmul2; Wproj: Conv1$\times$1(192$\to$192), BN. \\
  & \textit{SFFN} (Block1) 
  & Up: LIF + Conv1$\times$1(192$\to$768), BN; Conv: LIF + Conv3$\times$3(768$\to$768, groups=12) + BN; Down: LIF + Conv1$\times$1(768$\to$192), BN. \\
  & \textit{DSSA} 
  & LIF; Conv2d(192$\to$384, 2$\times$2, stride=2), BN; matmul1, matmul2; Wproj: Conv1$\times$1(192$\to$192), BN. \\
  & \textit{SFFN} (Block2) 
  & Up: LIF + Conv1$\times$1(192$\to$768), BN; Conv: LIF + Conv3$\times$3(768$\to$768, groups=12) + BN; Down: LIF + Conv1$\times$1(768$\to$192), BN. \\
\hline

\textbf{Stage (2)} 
  & \textit{Downsample} 
  & Conv3$\times$3(192$\to$384, stride=2), BN, LIF \\
  & \textit{SSDSA} 
  & LIF; Conv2d(384$\to$768), BN; matmul1, matmul2; Wproj(384$\to$384) + BN + SSDP (unsupervised plasticity). \\
  & \textit{SFFN} (Block1) 
  & Up: LIF + Conv1$\times$1(384$\to$1536), BN; Conv: LIF + Conv3$\times$3(1536$\to$1536, groups=24) + BN; Down: LIF + Conv1$\times$1(1536$\to$384), BN. \\
  & \textit{SSDSA} 
  & Similar spiking attention + Wproj + SSDP, (384$\to$768). \\
  & \textit{SFFN} (Block2) 
  & Channels (384$\leftrightarrow$1536), same pattern of Up/Conv/Down. \\
  & \textit{SSDSA} 
  & Repeated spiking attention with SSDP. \\
  & \textit{SFFN} (Block3) 
  & Similar final up/conv/down structure. \\
\hline

\textbf{AvgPool} 
  & AdaptiveAvgPool2d((1,1)) 
  & step\_mode=m, merges spatial dims. \\
\hline

\textbf{Classifier} 
  & Linear(in\_features=384 $\to$ out\_features) 
  & Output dimension for classification. \\
\hline

\textbf{SSDP Modules} 
  & \multicolumn{2}{p{0.82\linewidth}|}{(SSDP), (SSDSA) for unsupervised plasticity in certain SSDSA layers} \\
\hline
\end{tabular}
\end{adjustbox}
\end{table}

\section{Supplementary Note 5: Integrating SSDP into the Attention Mechanism of SNN-Transformer}
\label{sup5}

Let $x\in\mathbb{R}^{T\times B\times C\times H\times W}$ be the tensor cached inside a DSSA block, where $T$ is the number of time steps, $B$ the batch size, $C$ channels, and $(H,W)$ spatial size. In \texttt{DSSAWithSSDP} we cache both the block input and output, $x_{\mathrm{in}}$ and $x_{\mathrm{out}}$.

\paragraph{(1) Spatial collapse and spike trains.}
We reduce spatial dimensions by a logical AND:
\begin{equation}
    \tilde s_{\mathrm{in}}(t,b,c)=\max_{u,v}\mathbb{1}[x_{\mathrm{in}}(t,b,c,u,v)>0]\quad
\tilde s_{\mathrm{out}}(t,b,c)=\max_{u,v}\mathbb{1}[x_{\mathrm{out}}(t,b,c,u,v)>0]
\end{equation}

yielding spike trains $\tilde s_{\mathrm{in/out}}\in\{0,1\}^{T\times B\times C}$.

\paragraph{(2) First-spike times and sample-level indicators.}
The first-spike indices are
\begin{equation}
t_{\mathrm{in}}(b,c)=\min\{t:\tilde s_{\mathrm{in}}(t,b,c)=1\}\quad
t_{\mathrm{out}}(b,c)=\min\{t:\tilde s_{\mathrm{out}}(t,b,c)=1\}
\end{equation}
with the convention $\min\emptyset=T$ (no spike).
Define sample-level indicators
\begin{equation}
Q_{b,i}=\mathbb{1}\!\left[\sum_{t}\tilde s_{\mathrm{in}}(t,b,i)>0\right]\quad
P_{b,j}=\mathbb{1}\!\left[\sum_{t}\tilde s_{\mathrm{out}}(t,b,j)>0\right]
\end{equation}
and the synchrony gate (broadcast outer product)
\begin{equation}
\lambda_{b,ji}=P_{b,j}\,Q_{b,i}\in\{0,1\}.
\end{equation}

\paragraph{(3) Pairwise temporal distance and Gaussian weight.}
Broadcast the first-spike times to form
\begin{equation}
\Delta t_{b,ji}=\big|\,t_{\mathrm{out}}(b,j)-t_{\mathrm{in}}(b,i)\,\big|\ \in\ \mathbb{R}_{\ge0}^{C_{\mathrm{out}}\times C_{\mathrm{in}}}
\end{equation}
and the symmetric kernel
\begin{equation}
g_{b,ji}=\exp\!\Big(-\frac{\Delta t_{b,ji}^{2}}{2\sigma^{2}}\Big)
\end{equation}

\paragraph{(4) SSDP update per mini-batch (no grad, after the optimiser step).}
For the target weight matrix (either the classifier weight $W_{\mathrm{clf}}\in\mathbb{R}^{C_{\mathrm{out}}\times C_{\mathrm{in}}}$ or the DSSA $1{\times}1$ projection $W_{\mathrm{proj}}\in\mathbb{R}^{C_{\mathrm{out}}\times C_{\mathrm{in}}}$), the batch update is
\begin{equation}
\Delta W\;=\;\frac{1}{B}\sum_{b=1}^{B}\Big(A_{+}\,\lambda_{b}\;-\;A_{-}\,(1-\lambda_{b})\Big)\odot g_{b}
\qquad
W\leftarrow \mathrm{clip}\big(W+\Delta W\big)
\end{equation}
Here $\odot$ denotes elementwise multiplication. The computation uses only binary gates and $\Delta t$; complexity per batch is $\mathcal{O}(C_{\mathrm{out}}C_{\mathrm{in}})$. No state or compute is added at inference.

\paragraph{(5) Hook placement and scheduling.}
SSDP is applied at two hooks: (i) the classifier linear layer, and (ii) the $1{\times}1$ projection of the last DSSA block in the final stage. Updates are performed \emph{after} each optimiser step and only when the current epoch $e\ge e_{\mathrm{start}}$ (warm-up). All SSDP operations run under \texttt{no\_grad} and use detached activations, leaving back-prop gradients unchanged.

\begin{wrapfigure}{r}{0.55\textwidth} 
  \centering
  \includegraphics[width=0.52\textwidth]{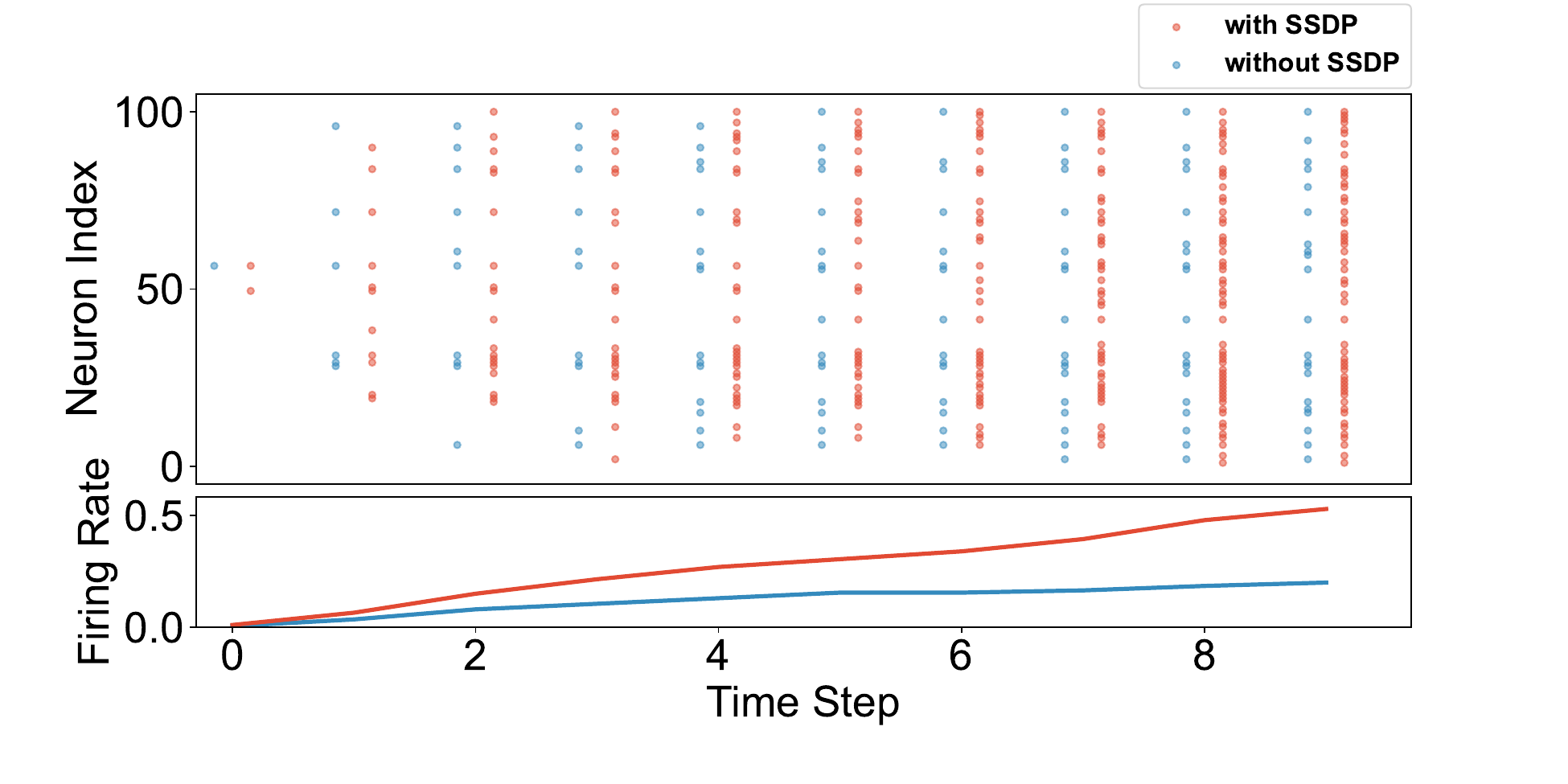}
  \caption{Synchrony events and average firing rate on Fashion-MNIST. 
  Top: spike rasters comparing SSDP (red) and no-SSDP (blue). 
  Bottom: average firing rate over time.}
  \label{fig:raster_fmnist}
\end{wrapfigure}

\section{Supplementary Note 6: Extended Evaluation of SSDP}
\label{sup6}

To supplement the main results presented in the manuscript, we provide a detailed performance comparison of the SNN-Transformer architecture on the CIFAR-100 dataset, evaluated with and without the proposed SSDP mechanism. In addition to accuracy metrics, we report throughput, multiply-accumulate operations (MACs \cite{han2015learning,fang2021deep}), synaptic operations (SOPs), and average power consumption.  These metrics, alongside average power consumption, offer a more comprehensive view of efficiency across architectures. Profiling was conducted using the integrated tools in SpikingJelly \cite{fang2023spikingjelly}. Table~\ref{tab:ssdp-cifar100-extended} summarizes the key results. 

The integration of SSDP led to a modest but meaningful increase in top-1 accuracy, from 78.73\% to 79.48\%, corresponding to an approximate 3.5\% reduction in classification error. Top-5 accuracy remained comparable, indicating that SSDP enhances first-choice decision precision without degrading overall output diversity. Importantly, this improvement was achieved with negligible impact on inference efficiency: throughput, MACs, and parameter count remained unchanged, while only minimal increases were observed in synaptic operations (SOPs; +2.2\%) and per-sample energy consumption (+2.1\%). Although the accuracy-to-spike power ratio (A/S Power Ratio) declined slightly from 11.43 to 11.19, the overall energy-performance trade-off remained highly favorable. These results highlight that SSDP offers consistent performance gains with minimal cost, reinforcing its practicality for neuromorphic applications where efficiency is critical.

\section{Supplementary Note 7: Additional qualitative rasters}
\label{sup7}

 Fig.~\ref{fig:raster_fmnist} shows example spike train visualizations from a subset of excitatory neurons in a one-layer SNN on the Fashion-MNIST task. In the non-SSDP trained network, spikes are distributed irregularly in time across neurons, with relatively few coincident events. In contrast, the SSDP-trained network displays clear episodes of synchronized firing: many neurons fire in unison, forming pronounced vertical bands in the raster plot. These synchronous events occur reliably in response to specific input patterns (e.g., particular digits), suggesting that SSDP causes ensembles of neurons to jointly encode those patterns. This supports the interpretation that SSDP’s learning rule reorganizes synaptic weights so that neurons that detect the same feature tend to fire together, effectively creating cell assemblies reminiscent of those hypothesized in vivo \cite{hebb2005organization,buzsaki2010neural}. Notably, the phenomenon was observed across network types and layers -- for instance, early convolutional layers of the spiking ResNet with SSDP also showed heightened coincident spiking compared to STDP, indicating that synchrony reinforcement is a general effect and not limited to shallow networks.

\bigskip

\end{document}